\def\year{2021}\relax
\documentclass[letterpaper]{article} 
\usepackage{aaai21}  

\usepackage[square,sort&compress,comma,numbers]{natbib}

\usepackage[utf8]{inputenc} 
\usepackage[T1]{fontenc}    
\usepackage{url}            
\usepackage{booktabs}       
\usepackage{amsfonts}       
\usepackage{nicefrac}       
\usepackage{microtype}      
\usepackage{soul}
\usepackage[utf8]{inputenc}
\usepackage{graphicx}
\usepackage{amsmath}
\usepackage{booktabs}
\usepackage{outlines}

\usepackage[colorlinks=true,
citecolor=blue,
filecolor=black,
linkcolor=blue,
urlcolor=blue]{hyperref}

\usepackage{microtype}
\usepackage{graphicx}

\usepackage{amsmath,amsfonts,bm}

\usepackage{xcolor}
\usepackage{colortbl}

\def\eqref#1{equation~\ref{#1}}

\def\mD{{\bm{D}}}

\DeclareMathAlphabet{\mathsfit}{\encodingdefault}{\sfdefault}{m}{sl}
\SetMathAlphabet{\mathsfit}{bold}{\encodingdefault}{\sfdefault}{bx}{n}

\def\sR{{\mathbb{R}}}

\newcommand{\E}{\mathbb{E}}

\usepackage{multirow}
\usepackage{paralist}

\usepackage{multicol}
\usepackage{diagbox}

\usepackage[ruled,noend]{algorithm2e}

\SetCommentSty{mycommfont}

\usepackage{here}

\usepackage{amsmath,amssymb,amsfonts,amsbsy,amsfonts,latexsym}
\usepackage{multirow}
\usepackage{makecell}
\usepackage[labelfont=bf,textfont=it,belowskip=0pt,aboveskip=5pt,tableposition=top]{caption}
\usepackage{xcolor}
\usepackage{colortbl}

\definecolor{colorA}{RGB}{189,201,225}
\definecolor{colorB}{RGB}{103,169,207}
\definecolor{colorC}{RGB}{ 28,144,153}
\definecolor{colorD}{RGB}{  1,108, 89}

\newcolumntype{R}{>{\columncolor{gray!40}}r}
\newcolumntype{L}{>{\columncolor{gray!40}}l}
\newcolumntype{C}{>{\columncolor{gray!40}}c}

\usepackage{tabularx,colortbl,xcolor}
\usepackage{multirow}
\usepackage[normalem]{ulem}
\useunder{\uline}{\ul}{}

\usepackage{enumitem}

\usepackage{xparse}

\DeclareGraphicsExtensions{.pdf,.png}

\SetKwInput{KwInput}{Input}
\SetKwInput{KwRequire}{Require}

\usepackage{pgfplots}

\usepackage{adjustbox}

\NewDocumentCommand{\var}{O{s} m O{}}{%
  \ensuremath{#1_{#2}^{#3}}
}
\usepackage{siunitx}

\newcommand{\commentout}[1]{}

\definecolor{light-gray}{gray}{0.80}

\newcommand\eref{Eq.~\ref}
\newcommand\fref{Figure~\ref}
\newcommand\tref{Table~\ref}
\newcommand\sref{Section~\ref}

\newcommand\ha{ \rowcolor{orange!0}}
\newcommand\hb{ \rowcolor{orange!15}}
\newcommand\hc{ \rowcolor{orange!40}}

\def\H{{\bf H}}
\def\g{{\bf g}}

\usepackage{amsthm}

\usepackage{amsthm}

\newtheorem{mytheorem}{Theorem}
\newtheorem{assumption}[mytheorem]{Assumption}
\newtheorem{lemma}[mytheorem]{Lemma}
\newtheorem{remk}[mytheorem]{Remark}

\def\loss{{\mathcal{L}}}

\def\Deltawt{{\Delta w_t}}
\def\wt{{w_t}}

\def\gt{{\g_t}}
\def\Ht{{\H_t}}

\usepackage[font=scriptsize]{subfig}

\newcommand{\recomsys}{RecSys\xspace}

\newcommand{\adam}{Adam\xspace}
\newcommand{\adamw}{AdamW\xspace}
\newcommand{\adagrad}{Adagrad\xspace}
\newcommand{\sgd}{SGD\xspace}
\newcommand{\OURS}{\textsc{AdaHessian}\xspace}

\usepackage{chngcntr}

\begin{document}

\title{\OURS: An Adaptive Second Order Optimizer for Machine Learning}

\author{
Zhewei Yao\thanks{Equal contribution.}$^{,1}$, Amir Gholami$^{*,1}$, Sheng Shen$^{1}$, Mustafa Mustafa$^{2}$, Kurt Keutzer$^{1}$, Michael W. Mahoney$^{1}$\\
$^{1}$University of California, Berkeley, $^{2}$NERSC, Lawrence Berkeley National Laboratory\\
{\tt\small \{zheweiy, amirgh, sheng.s, keutzer, mahoneymw\}@berkeley.edu}, \tt\small mmustafa@lbl.gov
}

\maketitle
\pagestyle{plain} 



\begin{abstract}
We introduce \OURS, a second order stochastic optimization algorithm  which
dynamically incorporates the curvature of the loss function via \textsc{ADA}ptive estimates of the \textsc{Hessian}.
Second order algorithms are among the most powerful optimization algorithms
with superior convergence properties as compared to first order methods such as \sgd and \adam.
The main disadvantage of traditional second order methods is their heavier per-iteration computation 
and poor accuracy as compared to first order methods.
To address these, we incorporate several novel approaches in \OURS, including:
(i) a fast Hutchinson based method to approximate the curvature matrix with low computational overhead;
(ii) a root-mean-square exponential moving average to smooth out variations of the Hessian diagonal across different iterations; and 
(iii) a block diagonal averaging to reduce the variance of Hessian diagonal elements.
We show that \OURS achieves new state-of-the-art results by a large margin as compared
to other adaptive optimization methods, including variants of \adam.
In particular, we perform extensive tests on CV, NLP, and recommendation system tasks and find that \OURS:
(i) achieves 1.80\%/1.45\% higher accuracy on ResNets20/32 on Cifar10, and 5.55\% higher accuracy on ImageNet as compared to \adam;
(ii) outperforms \adamw for transformers by 0.13/0.33 BLEU score on IWSLT14/WMT14 and 2.7/1.0 PPL on PTB/Wikitext-103; 
(iii) outperforms \adamw for SqueezeBert by 0.41 points on GLUE; 
and
(iv) achieves 0.032\% better score than \adagrad for DLRM on the Criteo Ad Kaggle dataset.
Importantly, we show that the cost per iteration of \OURS is comparable to first-order
methods, and that it exhibits robustness towards its hyperparameters.
The code for \OURS is open-sourced and publicly-available~\cite{adahessian}.
\end{abstract}  

\section{Introduction}
\label{sec:intro}

The high dimensional and non-convex nature of many machine learning tasks
has rendered many classical optimization methods inefficient for training and/or evaluating
Neural Network (NN) models. 
After decades of research, first order
methods, and in particular variants of Stochastic Gradient Descent (\sgd), have
become the main workhorse for training NN models. 
However, they are by no means an ideal solution for training NN models.
There are often a lot of ad-hoc rules that need to be followed very precisely
to converge (hopefully) to a point with good generalization properties.
Even the choice of the first order optimizer has become an ad-hoc rule which
can significantly affect the performance.
For example, \sgd with momentum is typically used in
Computer Vision (CV); \adam is used for training transformer models
for Natural Language Processing (NLP); and \adagrad is used
for Recommendation Systems (\recomsys). 
Using the wrong \sgd variant can lead to significant performance degradation.
Another challenging ad-hoc rule is the choice of hyperparameters and hyperparameter tuning methods,
even after an optimizer
is chosen. Hyperparameters include learning rate, decay schedule, choice
of momentum parameters, number of warmup iterations, etc.
As a result of these and other issues, one has to \emph{babysit} the optimizer
to make sure that training converges to an \emph{acceptable} training loss, without
any guarantee that a given number of iterations is enough to reach a local
minima.

Importantly, one may \emph{not} observe the above problems for certain popular learning tasks, such as ResNet50 training on ImageNet. 
The reason is that, for these tasks, years of industrial scale hyperparameter tuning has lead to what may be called \emph{ideal SGD behaviour}.
That is, for this problem, hyperparameters have been brute-force engineered to compensate for the deficiencies of first order methods.
Such a brute force approach is computationally and financially
not possible for many large scale learning problems---certainly it is not possible to do routinely---and this has made it challenging to train and apply NN models reliably.

\begin{figure}[t]
\begin{center}
  \includegraphics[width=.98\linewidth]{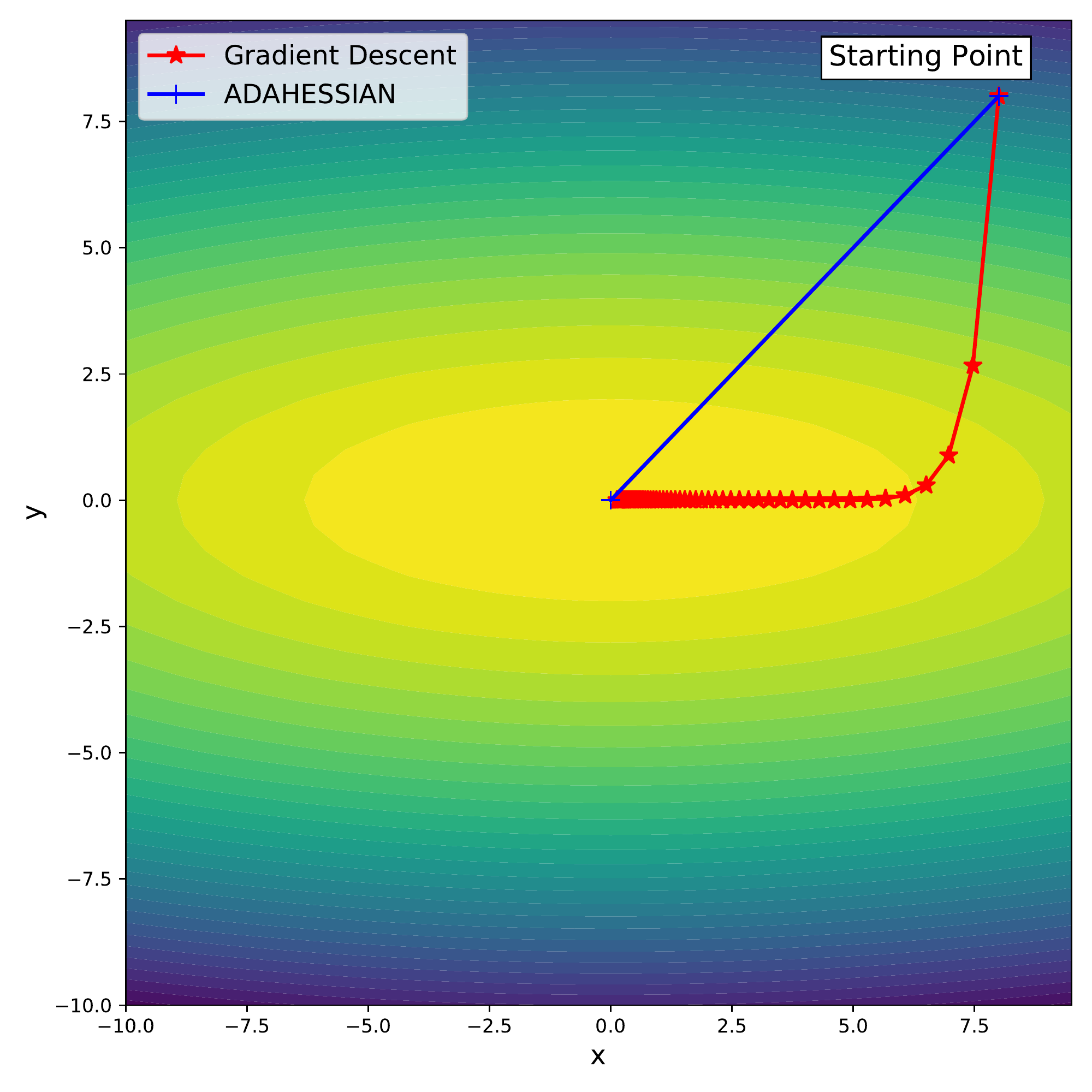}
\end{center}
 \caption{\footnotesize 
 The trajectory of gradient descent and \OURS on a simple 2D quadratic function $f(x, y) = 10x^2+y^2$. 
 Gradient descent converges very slowly, even though this problem has a reasonable condition number. 
 However, \OURS converges to the optimum in just one step. This is because second order methods
 normalize the curvature difference between x and y axis by preconditioning the gradient
 vector before the weight update (by rescaling and rotating the gradient vector).
}
\label{fig:quadratic_function}
\end{figure}

Many of these issues stem from the fact that first order methods only
use gradient information and do not consider the curvature properties of the loss landscape,
thereby leading to their suboptimal behaviour.
Second order methods, on the other hand, are specifically designed to capture and exploit the curvature of the loss
landscape and to incorporate both gradient and Hessian information. They are among the most
powerful optimization algorithms, and they have many favorable properties such as resiliency to \emph{ill-conditioned} loss landscapes, invariance
to parameter scaling, and robustness to hyperparameter tuning.
The main idea underlying second order methods involves \emph{preconditioning} the gradient vector
before using it for weight update. This has a very intuitive motivation related
to the curvature of the loss landscape.
For a general problem, different parameter dimensions exhibit different curvature properties. 
For example, the loss could be very flat in one dimension and very sharp in another. As a result,
the step size taken by the optimizer should be different for these dimensions, and we would 
prefer to take bigger steps for the flatter directions and relatively smaller steps for the 
sharper directions.
This can be illustrated with a simple 2D quadratic function as shown in~\fref{fig:quadratic_function},
where we show the trajectories of different optimizers.
As can be seen, first order methods need a large number of steps for convergence and/or are hard to converge at all without hyperparameter tuning. 
However, second order methods capture this curvature difference, by normalizing
different dimensions through rotation and scaling of the gradient vector before the weight update. Nonetheless, this comes at a cost.
Despite the theoretically faster convergence rate of second order methods, they
are rarely used for training NN models. This is due in part to their high
computational cost. 

In this paper, however, we will show that it is possible
to compute \emph{approximately} an exponential moving average of the Hessian and use
it to precondition the gradient adaptively. The result is \OURS, an adaptive optimizer that
exceeds the state-of-the-art performance for a wide range of learning
problems, including ResNets~\cite{he2016deep} for CV, transformers~\cite{ott2018scaling} for NLP problems, and DLRM~\cite{naumov2019deep} models for \recomsys tasks. 
In more detail, the main contributions of our work are the~following.

\begin{outline}[enumerate]
    \1 To reduce the overhead of second order methods, we approximate the Hessian
    as a diagonal operator. This is achieved by applying Hutchinson's method
    to approximate the diagonal of the Hessian. Importantly, this approximation
    allows us to efficiently apply a root-mean-square exponential moving average to
    smooth out ``rugged'' loss surfaces. The advantage of this approach
    is that it has $\mathcal{O}(d)$ memory complexity.

    \1 We incorporate a block diagonal averaging to reduce the variance
    of Hessian diagonal elements. In particular, this has no additional
    computational overhead in the Hutchinson's method, but it favorably
    affects the performance of the optimizer.

    \1 To reduce
    \OURS overhead, we measure the sensitivity of \OURS to different hyperparameters such as learning 
    rate, block diagonal averaging size, and delayed Hessian computation. 
    Interestingly, our results show that \OURS is robust to those hyperparameters. 
    See~\sref{sec:blocksize_lr_effect} and~\ref{sec:result_delay_hessian_update}.

    \1 We extensively test \OURS on a wide range of learning tasks. In all tests,
    \OURS significantly outperforms other adaptive optimization methods. 
     Importantly note that these results
    are achieved even though  we use the same
    learning rate schedule, weight decay, warmup schedule, dropout, as well as first/second order moment coefficients.
    In particular, we consider the following tasks.
    
    \2 \textbf{Computer Vision:} \OURS achieves significantly higher accuracy,
    as compared to \adam. For instance, for ResNet32 on Cifar10, 
    \OURS achieves 93.08\% as opposed to 91.63\% achieved by \adam.
    Furthermore, for ResNet18 on ImageNet, \OURS achieves 70.08\% accuracy as opposed
    to 64.53\% of \adam. For all tests, \OURS achieves similar performance
    to the ideal SGD behavior, which is a result of hyperparameters having been tuned for 
    many years at the industrial scale.
    Comparison with other optimizers and other models
    is discussed in~\sref{sec:result_image_classification}.

    \2 \textbf{Natural Language Processing:}
    \OURS improves the performance of transformers for machine translation and language modeling tasks, as compared to \adamw.
    In particular, \OURS significantly outperforms \adamw by 0.13/0.33 BLEU on IWSLT14/WMT14, and by 2.7/1.0 PPL on PTB/WikiText-103. 
    Moreover, for SqueezeBERT~\cite{iandola2016squeezenet} fine-tuning on GLUE, \OURS achieves 0.41 better points than~\adamw. 
    See~\sref{sec:result_machine_translation}, \ref{sec:language_modeling}, and~\ref{sec:natrual_language_understading} for more details.
    \2 \textbf{Recommendation System:} \OURS improves the
    performance of DLRM on the Criteo Ad Kaggle dataset by 0.032\% as compared to \adagrad, which
    is commonly used. See~\sref{sec:recommendation_system} for more details. 
    \1 We measure the sensitivity of \OURS to different hyperparameters such as learning 
    rate, spatial averaging size, and delayed Hessian computation. 
    Interestingly, our results show that \OURS is robust to those hyperparameters. 
    See~\sref{sec:blocksize_lr_effect} and~\ref{sec:result_delay_hessian_update} for more details.
\end{outline}
We emphasize that our empirical results are achieved even though we use the same learning rate schedule, weight decay, warmup schedule, dropout, batch size, and first/second order moment coefficients as the heavily-tuned default first order baseline optimizers.
Additional gains could be achieved if one wanted to optimize extensively these hyperparameters.

\section{Problem Formulation and Related work}
\label{sec:related_work}

We focus on supervised learning tasks where the goal is to solve
a non-convex stochastic optimization problem of the form:
\begin{equation}\label{eq:basic_problem}
    \min_{\theta}\loss(\theta) = \frac1N\sum_{i=1}^N l_i(x_i, y_i; \theta),
\end{equation}
where $\theta\in\sR^d$ denotes the model parameters, $l_i(x_i, y_i; \theta)$ is the loss function, $(x_i,y_i)$ is the paired input data and its corresponding ground truth label, and
$N$ is the total number of data points in the training dataset. 
Furthermore, we denote the gradient of the loss w.r.t. model parameters as $\g=\frac{1}{N_B}\sum_{i=1}^{N_B}\frac{\partial l_i}{\partial \theta}$, and the corresponding
second derivative (i.e., Hessian) as $\H=\frac{1}{N_B}\sum_{i=1}^{N_B}\frac{\partial^2 l_i}{\partial \theta^2}$, where $N_B$ is the size of one mini-batch.

Solving~\eref{eq:basic_problem} for a real learning problem (and not a simple model) is a very challenging task. Despite years
of research, we have not yet been able to resolve several seemingly ad-hoc tricks
that are required to converge (hopefully) to a \emph{good} solution.
Next, we briefly discuss the different popular optimization methods proposed in recent years
to address the challenges associated with solving~\eref{eq:basic_problem}.
This is by no means a comprehensive review, and we refer the interested reader
to~\cite{bottou2018optimization} for a thorough review.

\subsection{Adaptive First Order Methods}

Due to their simplicity and effectiveness, 
first order optimization methods~\cite{robbins1951stochastic,nesterov1983method,duchi2011adaptive,zeiler2012adadelta,kingma2014adam,loshchilov2017decoupled} have become the de-facto algorithms used in deep learning.
There are multiple variations, but these methods can be represented using the following general update formula:
\begin{equation}
    \theta_{t+1} = \theta_t - \eta_t m_t / v_t,
    \label{e:sgd}
\end{equation}
where $\eta_t$ is the learning rate, and $m_t,\text{ and } v_t$ denote the so called first and second moment terms, respectively.
A simple and popular update method is SGD, originally proposed in 1951 as a root-solving algorithm~\cite{robbins1951stochastic}:
\begin{equation}
    m_t = \beta m_{t-1} + (1-\beta)\g_t~~~\text{and}~~~v_t \equiv 1.
\end{equation}
Here, $\g_t$ is the gradient of a mini-batch at $t$-th iteration and $\beta$ is the momentum hyperparameter.

Using \sgd to solve~\eref{eq:basic_problem} is often very challenging, as the convergence of the iterative formulae in~\eref{e:sgd}
is very sensitive to the right choice of the learning rate, its decay schedule, and the momentum parameter. 
To address this, several methods have been proposed to take into account the knowledge of the geometry of the
data by scaling gradient coordinates, using the past gradient information.
This can be viewed in one of two equivalent ways: either as automatically adjusting the 
learning rate in~\eref{e:sgd}; or as an adaptive \emph{preconditioning} of the gradient.
One notable method is \adagrad~\cite{duchi2011adaptive,mcmahan2010adaptive}, which accumulates
all the gradients from the first iteration and applies the square root of the result to
precondition the current gradient. The update formulae in this case become\footnote{Throughout the paper, without further notification, for two vectors, e.g., $a$ and $b$, we use both $ab$ and $a\odot b$ to denote the element-wise product, and $\langle a,b \rangle$ denotes the inner product.}:
\begin{equation}
\label{e:adagrad}
    m_t = \g_t~~~~~~\text{and}~~~~~v_t = \sqrt{\sum_{i=1}^t\g_i\g_i}.
\end{equation}

While \adagrad works well for sparse settings, its performance significantly degrades for dense settings, which
is the case for many machine learning tasks. 
In particular, this stems from the accumulation of all previous gradients for the preconditioner~\eref{e:adagrad}.
This results in a monotonic increase in the magnitude of the second moment, $v_t$, which effectively translates
into a rapid decay of the learning rate.
To address this, several methods have been proposed where the intuition is to limit the accumulation
to a small window of past iterations, and in particular exponentially reduce the weight of earlier iterations.
Notable works incorporating this method are RMSProp, ADADelta, and \adam~\cite{tieleman2012lecture,zeiler2012adadelta,kingma2014adam}.
In particular, for \adam~\cite{kingma2014adam}, the two moments for the update rule are the following:
\begin{equation}
\begin{split}
    m_t = \frac{(1-\beta_1)\sum_{i=1}^t\beta_1^{t-i}\g_i }{1-\beta_1^t}, \\
    v_t = \sqrt{\frac{(1-\beta_2)\sum_{i=1}^t \beta_2^{t-i}\g_i\g_i}{1-\beta_2^t}},
\end{split}
\end{equation}
where $0<\beta_1,~\beta_2<1$ are two hyperparameters sometimes referred to as first and second moment coefficients. 
In particular, note that the sum over past gradients is scaled by $\beta_2$ which exponentially reduces
the contribution of early gradients.
A summary of the different $m_t$ and $v_t$ used by common first-order optimizers is given in~\tref{tab:optimizer_summary}.
A notable variant here is \adamw~\cite{loshchilov2017decoupled}, which shows that decoupling weight decay from the update equation of \adam can lead to a noticeable performance improvement.
Recently, \adamw has become the preferred optimizer for NLP tasks, and in particular for training 
transformers~\cite{vaswani2017attention}. 
There are also many other variants of adaptive first order methods~\cite{chaudhari2019entropy,zhang2015deep,loshchilov2016sgdr,shazeer2018adafactor,singh2015layer}. 

Despite all these attempts, it is still not clear which optimizer should work for a \emph{new} learning task/model. 
This is in fact one of the main baffling practical issues in machine learning, and one for which theory has little to say. 
For example, \sgd is currently the best performing optimizer for some CV tasks. 
That is, using other variants such as \adamw leads to significantly worse generalization performance. 
However, for NLP tasks, \adamw has the best performance by a large margin as compared to \sgd.
The point here is that even the choice of the optimizer has effectively become a hyperparameter.

\begin{table}[t] 
\caption{\footnotesize 
Summary of the first and second moments used in different optimization
algorithms for updating model parameters ($\theta_{t+1} = \theta_t - \eta m_t / v_t$). 
Here $\beta_1$ and $\beta_2$ are first and second moment hyperparameters.
}
\label{tab:optimizer_summary}
\centering
\setlength\tabcolsep{2.pt}
\begin{adjustbox}{width=1\linewidth} 
\begin{tabular}{lcccccc}
\toprule
\ha Optimizer & $m_t$  & $v_t$   \\ 
\midrule
\ha \sgd~\cite{robbins1951stochastic}   & $\beta_1 m_{t-1} + (1-\beta_1)\g_t $  & 1 \\[8pt]
\ha \adagrad~\cite{duchi2011adaptive}   & $\g_t$                            & $\sqrt{\sum_{i=1}^t \g_i\g_i}$ \\[8pt]
\ha \adam~\cite{kingma2014adam}         & $\frac{(1-\beta_1)\sum_{i=1}^t\beta_1^{t-i}\g_i }{1-\beta_1^t}$   & $\sqrt{\frac{(1-\beta_2)\sum_{i=1}^t \beta_2^{t-i}\g_i\g_i}{1-\beta_2^t}}$ \\[8pt]
\ha RMSProp~\cite{tieleman2012lecture}  & $\g_t$                                                            & $\sqrt{\beta_2 v_{t-1}^2 + (1-\beta_2)\g_t\g_t}$ \\[8pt]
\midrule
\hc \OURS                               & $\frac{(1-\beta_1)\sum_{i=1}^t\beta_1^{t-i}\g_i }{1-\beta_1^t}$   & $\sqrt{\frac{(1-\beta_2)\sum_{i=1}^t \beta_2^{t-i}\mD_i^{(s)}\mD_i^{(s)}}{1-\beta_2^t}}$ \\
\bottomrule 
\end{tabular}
\end{adjustbox}
\end{table}

\subsection{Second Order Methods}

Second order methods are among the most powerful optimization methods that have been designed, and
there have been several attempts to use their many advantages for training NNs.
Second order methods are designed to address \emph{ill-conditioned} optimization problems
by automatically rotating and rescaling the gradient. This allows one to choose a better descent direction, and to adjust automatically the learning rate for each parameter. 
There have also been multiple theoretical studies showing better convergence
rate of second order based methods~\cite{bollapragada2019exact,yao2018inexact,roosta2016sub,xu2016sub,xu2017newton,xu2017second,conn2000trust,agarwal2016second,wang2018giant,agarwal2016finding,carmon2018accelerated,chen2019fast}.
In particular, second order methods can guarantee convergence to
second order critical points, while the vast majority
of first order methods lack such guarantees. 
For example, theoretically it has
been shown that
some first order methods can only converge to an approximate second order critical point~\cite{ge2015escaping,jin2017escape,levy2016power,reddi2017generic}.

\begin{figure*}[t]
\begin{center}
  \includegraphics[width=.90\linewidth]{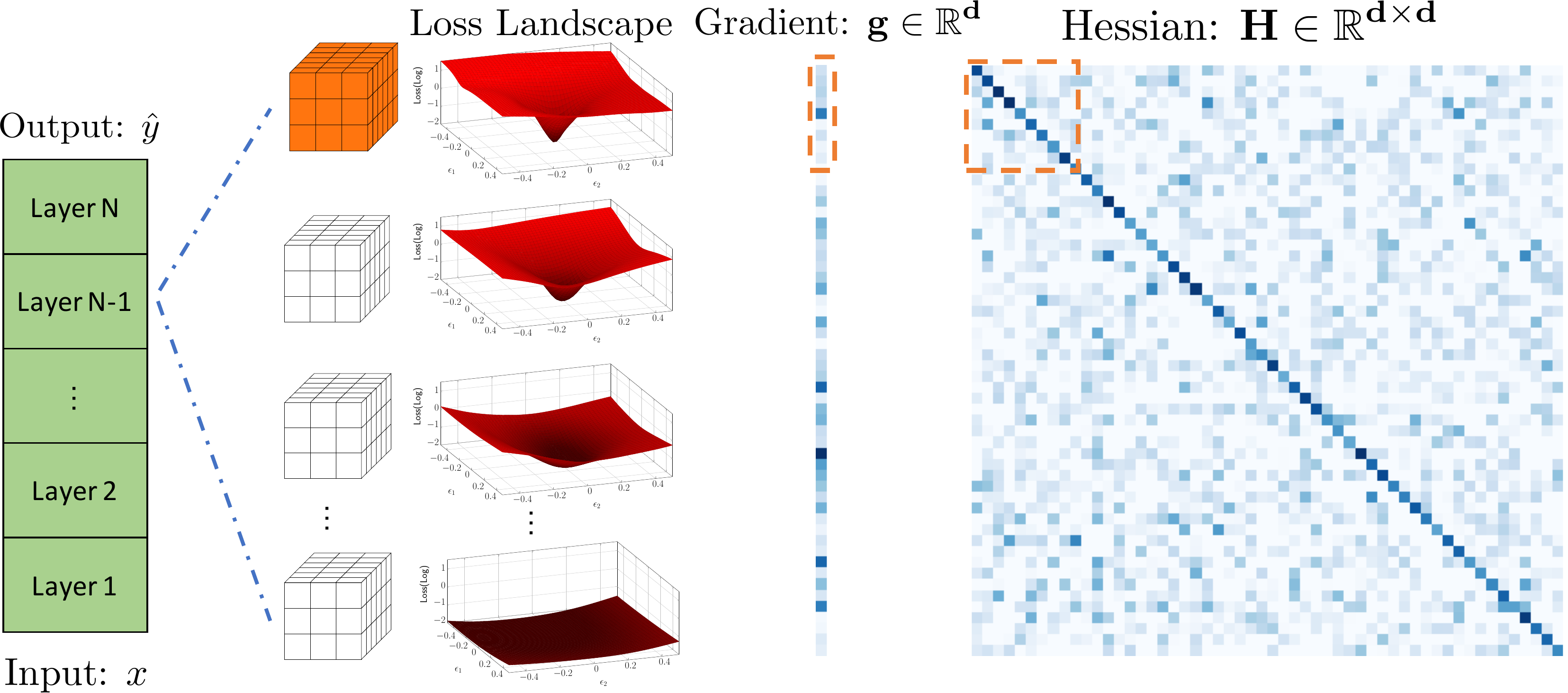}
\end{center}
 \caption{\footnotesize 
A simple model with N layers (first column); with the convolutional blocks of the N-1 layer shown (second column); and the loss landscape of each block (third column), which can be calculated by perturbing the convolutions's parameters
in two different eigendirections. 
(See~\cite{yao2019pyhessian} for details of how to construct loss landscape.)
Note the different loss landscape topologies. 
First order methods do not explicitly capture this difference. 
The entries (3D tensors) colored in orange show the components used for calculating the spatial average of Hessian.
The part of the gradient (fourth panel) highlighted in the orange box is the corresponding gradient of the orange convolution kernel; and the part of the Hessian diagonal (fifth panel) highlighted in the orange box is used to compute the spatial average.
}
\label{fig:block_hessian_vis}
\end{figure*}

Newton's method is a classical second order method where one solves a linear system, essentially meaning that the inverse of the local Hessian is used at every
iteration to precondition the gradient vector.%
\footnote{To be clear, when we refer to computing an inverse, we mean that we use a numerical method that performs a linear equation solve that effectively amounts to working with the inverse implicitly.  Of course, one would never actually compute the Hessian inverse explicitly.}
One major challenge with this
approach is that it can be expensive to solve the linear system, na\"{\i}vely requiring cubic computational complexity, not including the cost of forming the Hessian itself and
the corresponding quadratic memory complexity.
However, the overhead of such a na\"{\i}ve implementation can be improved by
using so-called matrix free methods, where the Hessian matrix is never explicitly formed (addressing quadratic
memory cost), and its inverse is approximately and only implicitly applied (addressing the cubic computational complexity).

One seminal work here is the limited memory BFGS (LBFGS)~\cite{byrd1995limited} method which has
a desirable linear computational and memory complexity. This approach approximates the
Hessian as a series sum of first order information from prior iterations. As such, these
approaches that do not directly use the Hessian operator are referred to as \emph{Quasi-Newton} methods.
While this approach works well for many optimization problems, it does not work
well for many machine learning problems. One reason for this is that LBFGS method
requires full batch gradients, as stochastic gradients can lead to drastic
errors in the approximation~\cite{bollapragada2018progressive}. This is one
of the main challenges with Quasi-Newton methods applied to machine learning problems~\cite{aghazadeh2020bear}.
Other approaches such as approximating the Hessian as the Kronecker product of vectors
have also been explored~\cite{martens2015optimizing}.

There has also been work on enhancing first order methods by incorporating the Fisher
information matrix~\cite{gupta2018shampoo}. The main idea is to use the Fisher
information instead of the squared norm of the gradient. A na\"{\i}ve use of Fisher information
has computational and memory overhead, but it is possible to also approximate the Fisher
information matrix using low rank decomposition~\cite{gupta2018shampoo}.

Another line of work has been to incorporate automatically the Hessian operator itself, instead
of approximating it using first order information. A 
work here
is~\cite{schaul2013no} which uses Gauss-Newton Hessian diagonal to 
adjust adaptively the learning rate. 
The work of~\cite{xu2017second} also directly incorporates
the Hessian using a trust region~method.

While the above approaches are very interesting
and result in a good performance for simple models, they do not achieve
comparable results for more complex NN architectures.
One of the reasons that second order methods have not been successful yet for machine learning,
as opposed to other domains such as scientific computing, is due to the stochastic
nature of the problem. Such stochastic noise leads to an erroneous approximation of the Hessian, leading
to suboptimal descent directions. 
SGD is more robust to such noise since we can efficiently incorporate moving averages and momentum.
Ideally, if there was a way to apply the same moving average method to the Hessian, then that would help smooth out local curvature noise to get a better approximation to the non-noisy
curvature of the loss landscape. However, such an approximation is challenging
since the Hessian is a matrix that cannot be explicitly formed to be averaged, whereas 
it is easy to form the gradient vector.

As we show below, \OURS addresses this problem by incorporating the Hutchinson's method
along with spatial averaging to reduce the impact of the stochastic noise. The result exceeds
the performance of all the above methods for machine learning tasks. 
Next, we formally introduce the \OURS algorithm.

\begin{figure}[t]
\begin{center}
  \includegraphics[width=.98\linewidth]{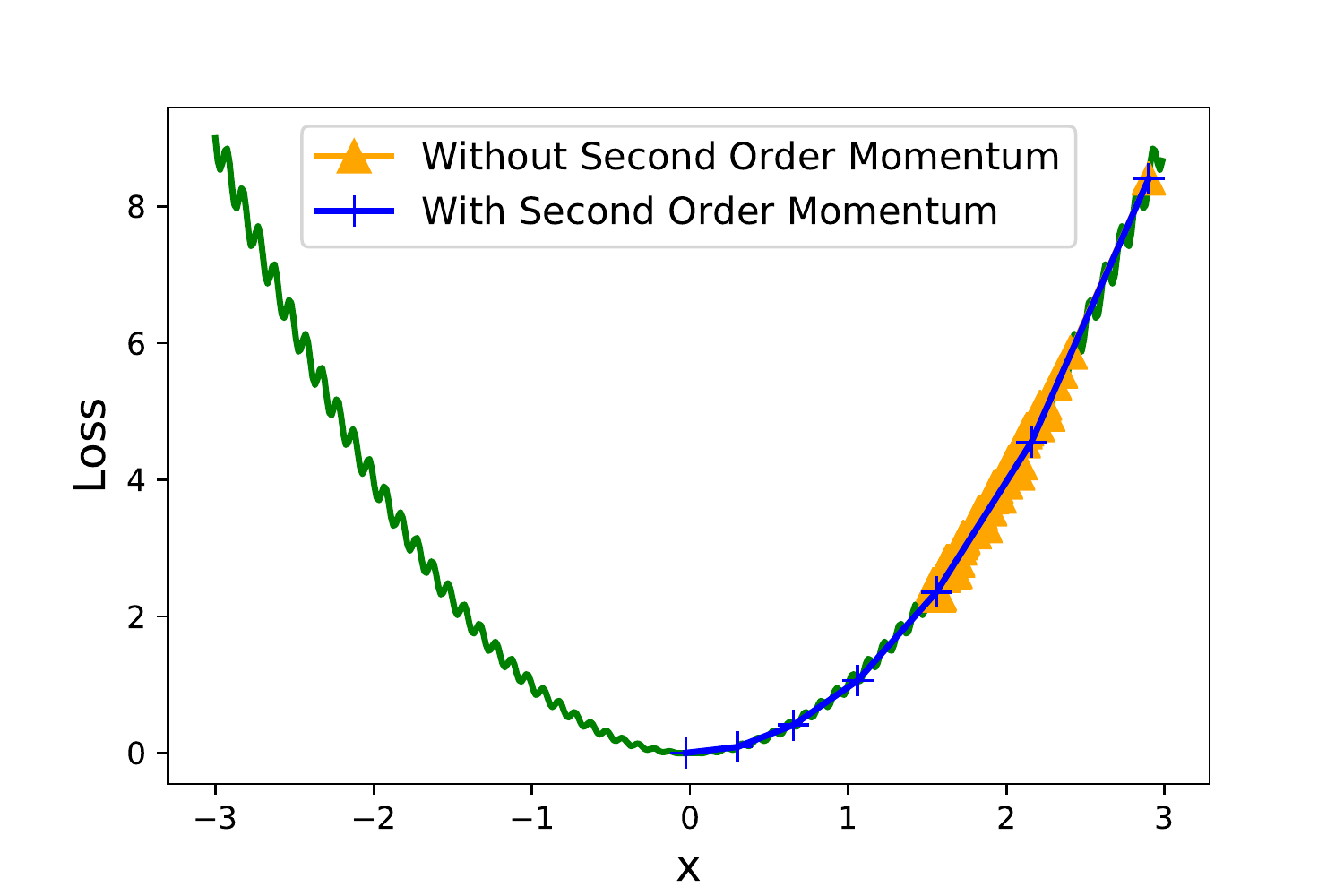}
\end{center}
 \caption{\footnotesize
 Local versus global curvature.
 Illustration of the local  curvature which can be noisy, and the global curvature with a
 simple 1D problem $f(x) = x^2 + 0.1x\sin(20\pi x)$.
 Using the exponential moving average of~\eref{eq:update_gh_adahess} is key to avoid
 the misleading local curvature information.
 To demonstrate this we test \OURS without moving average (orange trajectory) which does
 not converge even after 1000 iterations. On the other hand, \OURS converges in
 7 iterations with the moving average enabled.
 }
\label{fig:2nd_momentum_illustration}
\end{figure}

\section{Methodological Approach}
\label{sec:methodology}
Here, we first provide the formulation for the full Newton method in~\sref{sec:adahessian}.
Then, we describe the three components of \OURS, namely Hessian diagonal approximation (\sref{sec:diagonal_hessian_approximation}), spatial averaging (\sref{sec:spatial_averaging}), and Hessian momentum (\sref{sec:hessian_momentum}).
Finally, we discuss the overall formulation of \OURS in~\sref{sec:adahessian_deeplearning}.

\subsection{A General Hessian Based Descent Direction}
\label{sec:adahessian}

For the loss function $f(w): \sR^d \rightarrow \sR$, let us denote the corresponding gradient and Hessian of $f(w_t)$ at iteration $t$ as $\g_t$, and $\H_t$, respectively.%
\footnote{Without confusion, we use the same gradient and Hessian notations for $f(w)$ and $\loss(\theta)$. Furthermore, when there is no confusion we will drop subscript $t$.} 
A general descent direction can then be written as follows for a positive-definite Hessian:
\begin{equation}
\small
\label{eq:gradient_hessian_direction}
    \Delta w_t = \H_t^{-k}\g_t,~~~\text{where}~~\H_t^{-k} = U_t^T\Lambda_t^{-k}U_t.
\end{equation}
Here, we refer to $0\leq k\leq 1$ as \emph{Hessian power}, and $U_t^T\Lambda_t U_t$ is the eigendecomposition of $\H_t$.
Note that for $k=0$, we recover the gradient descent method; and for $k=1$, we recover the Newton method.
In our empirical tests we consider non-convex machine learning problems, but we provide a standard
convergence behaviour of~\eref{eq:gradient_hessian_direction} in Appendix~\ref{sec:proof_gfh_direction} for a simple  strongly convex and strictly smooth function $f(w)$. (We emphasize that the proof is very standard and we are only
including it for completeness.)

The basic idea of Hessian based methods is to \emph{precondition} the gradient with the $\H^{-k}$ and use $\H^{-k}\g$ for the update direction, instead of using the \emph{bare} gradient $\g$ vector.
The preconditioner automatically rotates and rescales the gradient vector.
This is important since the loss landscape curvature is generally different across different directions/layers and since these directions need not correspond to the canonical axes.
This is illustrated in~\fref{fig:block_hessian_vis}, where we show a 2D schematic plot of the loss landscape for different convolution channels~\cite{yao2019pyhessian}. 
Each channel can have a different loss landscape topology. For example, the last channel has a much flatter loss landscape, as compared to other layers. 
As a result, it is preferable to take a larger step size for the last channel than for the first channel, which has a very ``sharp'' loss landscape.
Problems that exhibit this behaviour are \emph{ill-conditioned}.
The role of the Hessian is to automatically normalize this ill-conditionedness by stretching and contracting different directions to accommodate for the curvature differences (full Newton method also rotates the gradient vector along with adjusting the
step size).

However, there are two major problems with this approach. 
The first problem is that a na\"{\i}ve use of the Hessian preconditioner comes at the prohibitively high cost of applying Hessian inverse to the
gradient vector at every iteration ($\H^{-k}\g$ term). 
The second and more challenging problem is that local Hessian (curvature)
information can be very misleading for a noisy loss landscape.
A simple example is illustrated in~\fref{fig:2nd_momentum_illustration}, where we plot a simple parabola with a small 
sinusoidal noise as the loss landscape (shown in green). As one can see, the local Hessian 
(curvature) information is completely misleading, as it computes the curvature of the sinusoidal
noise instead of global Hessian information for the parabola. Applying such misleading 
information as the preconditioner would actually result in very small steps to converge to one 
of the many local minima created by the sinusoidal noise.
The same problem exists for the gradient as well, but that can be alleviated
by using gradient momentum instead of local gradient information. 
However, as mentioned before it is computationally
infeasible
to compute (na\"{\i}vely) a Hessian momentum. 
The reason is that we cannot form the Hessian matrix and
average it throughout different iterations, as such an approach has quadratic memory complexity
in the number of parameters along with a prohibitive computational cost.
However, one could use Randomized Numerical Linear Algebra to get a sketch of the Hessian matrix~\cite{yao2018hessian,yao2019pyhessian,gupta2019oversketched}.
In particular, we show how this can be done to approximate the Hessian diagonal.
However, as we discuss next, both problems can be resolved by using Hessian diagonal instead of the full Hessian.

\subsection{Hessian Diagonal Approximation}
\label{sec:diagonal_hessian_approximation}
To address the issue that applying the inverse Hessian to the gradient vector at every iteration is computationally infeasible, one could use an inexact Newton method, where an approximate Hessian operator is used instead of the full Hessian~\cite{dembo1982inexact,xuNonconvexTheoretical2017,xu2017newton,yao2018inexact,bollapragada2019exact}.
The most simple and computationally efficient approach is to approximate the Hessian as a diagonal operator in~\eref{eq:gradient_hessian_direction}:
\begin{equation}
\small
\label{eq:gradient_diagonal_direction}
    \Delta w = diag(\H)^{-k}\g,
\end{equation}
where $diag(\H)$ is the Hessian diagonal, which we denote as $\mD$.%
\footnote{Note that $\mD$ can be viewed as a vector, in which case $\mD^{-k}\g$ is an element-wise product of vectors. Without clarification, $\mD$ is treated as a vector for the rest of the paper.}  
We show that using~\eref{eq:gradient_diagonal_direction}
has the same convergence rate as using~\eref{eq:gradient_hessian_direction} for 
simple strongly convex and strictly smooth function $f(w)$ (see Appendix~\ref{sec:proof_gdh_direction}). 
Note that we only include the proof for completeness, and our algorithm \OURS can be applied for general machine learning problems. 

\begin{figure}
  \centering
    \centering
    \raisebox{-\height}{\includegraphics[width=.99\linewidth]{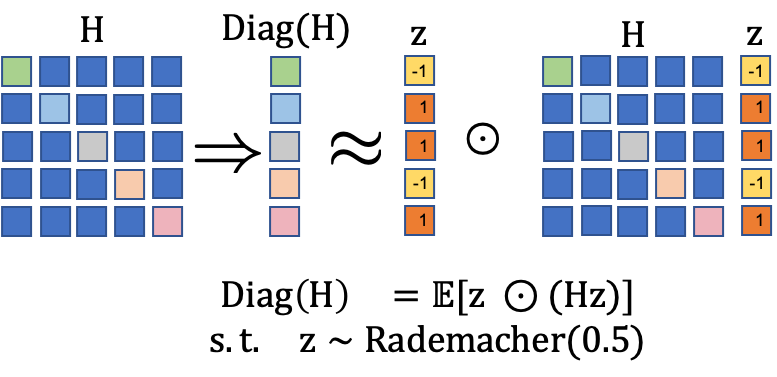}}
    \caption{\footnotesize 
    Illustration of the diagonal Hessian estimation
    with Hutchinson's method. 
    }
    \label{fig:diag_computation}
\end{figure}

The Hessian diagonal $\mD$ can be efficiently computed using the Hutchinson's method.
The two techniques we use for this approximation are: (i) a Hessian-free method~\cite{yao2018hessian}; and (ii) a randomized numerical linear algebra (RandNLA) method~\cite[Figure 1]{bekas2007estimator}. 
In particular, the Hessian-free method is an oracle to compute the multiplication between the Hessian matrix $\H$ with a random vector $z$, i.e., 
\begin{equation}
\label{eq:hessian_matvec}
\small
    \frac{\partial \g^Tz}{\partial \theta} = \frac{\partial \g^T}{\partial \theta}z + \g^T\frac{\partial z}{\partial \theta} = \frac{\partial \g^T}{\partial \theta}z = \H z.
\end{equation}
Here, the first equality is the chain rule, and the second equality is since $z$ is independent of $\theta$.
\eref{eq:hessian_matvec} effectively allows us to compute the Hessian times a vector $z$, without having to form explicitly the Hessian, by backpropotating the $\g^Tz$ term.
This has the same cost as ordinary gradient backpropogation~\cite{yao2018hessian}.
Then, with the Hessian matvec oracle, one can compute the Hessian diagonal using Hutchinson's method:
\begin{equation}
\small
\label{eq:diag_estimation}
    \mD = diag(\H) = \E[z\odot(\H z)],
\end{equation}
where $z$ is a random vector with Rademacher distribution,
and $\H z$ is computed by the Hessian matvec oracle given in~\eref{eq:hessian_matvec}. 
This process is illustrated in ~\fref{fig:diag_computation}.
It can be proved that the expectation of $z\odot(\H z)$ is the Hessian diagonal~\cite{bekas2007estimator}.

\begin{figure*}[!htbp]
\begin{center}
  \includegraphics[width=0.9\linewidth]{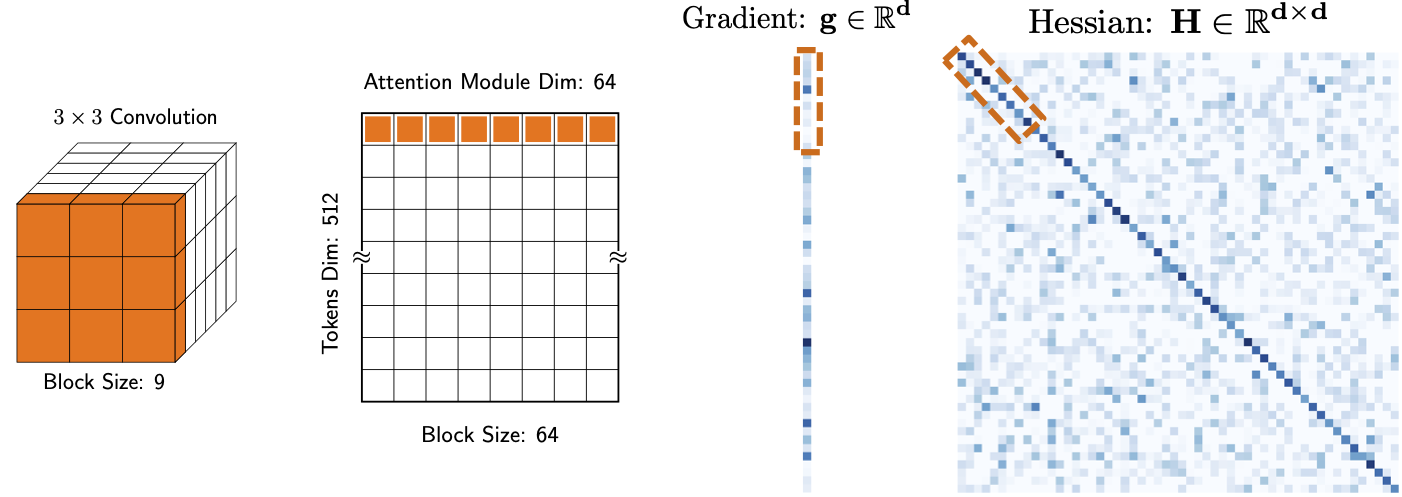}
\end{center}
 \caption{\footnotesize 
 Illustration of the block size used to average the Hessian diagonal to
 smooth spatial variations. For a convolution layer, we average
 each channel (groups of 9 parameters); and for multi-head
 attention, we average consecutive elements along the rows (attention dimension). We found that using block averaging helps, although
 \OURS is not very sensitive to this hyperparameter as illustrated in~\tref{tab:block_stable_iwslt}.
 }
\label{fig:block_hessian_vis_conv_multihead}
\end{figure*}

Another important advantage, besides computational efficiency, of using the Hessian diagonal is that we can compute its moving average to resolve the local noisy Hessian as mentioned at the end of~\sref{sec:adahessian}. 
This allows us to smooth out noisy local curvature information, and to obtain estimates that use global Hessian information instead.
We incorporate both spatial averaging and momentum (temporal averaging) to smooth out this noisy Hessian estimate as described next.

\subsection{Spatial Averaging}
\label{sec:spatial_averaging}

The Hessian diagonal can vary significantly for each single parameter dimension of the problem. 
We found it helpful to perform spatial averaging of Hessian diagonal and use
the average to smooth out spatial variations.
For example, for a convolutional layer, each convolution parameter can have a
very different Hessian diagonal. In \OURS we compute the average of the Hessian diagonal
for each convolution kernel ($3\times3$) as illustrated in~\fref{fig:block_hessian_vis_conv_multihead}.
Mathematically, we perform a simple spatial averaging on the Hessian diagonal as follows:
\begin{equation}
\label{eq:spatial_averaging}
\small
    \mD^{(s)}[ib + j] = \frac{\sum_{k=1}^{b}\mD[ib+k]}{b},\text{for~} 1\leq j\leq b,0\leq i\leq \frac{d}{b}-1,
\end{equation}
where $\mD \in \sR^d$ is the Hessian diagonal, $\mD^{(s)}\in\sR^d$ is the spatially
averaged Hessian diagonal, $\mD[i]$ ($\mD^{(s)}[i]$) refers to the i-th element of $\mD$ ($\mD^{(s)}$), $b$ is the spatial average block size, and $d$ is the number of model parameters divisible by $b$. 
We show that replacing $\mD$ in~\eref{eq:gradient_diagonal_direction} by $\mD^{(s)}$ in~\eref{eq:spatial_averaging},
the update direction has the same convergence rate as using~\eref{eq:gradient_hessian_direction} for 
simple strongly convex and strictly smooth function $f(w)$ (see Appendix~\ref{sec:proof_spatial_averaging}). 
Note that we only include the proof for completeness, and our algorithm \OURS can be applied for general machine learning problems.

Figure~\ref{fig:block_hessian_vis_conv_multihead} provides illustration of spatial averaging for both convolutional and matrix kernels. 
In general, the block size $b$ is a hyperparameter that can be tuned for different tasks.
While this is a new hyperparameter that can help the performance,
the performance of \OURS is not very sensitive to it (we provide sensitivity results
in~\sref{sec:blocksize_lr_effect}).

Next we describe momentum which is another useful method to smooth out Hessian noise over different iterations.

\begin{algorithm}
\caption{\OURS}
\label{alg:adahess}
\SetAlgoLined
    \KwRequire{Initial Parameter: $\theta_0$}
    \KwRequire{Learning rate: $\eta$}
    \KwRequire{Exponential decay rates: $\beta_1$, $\beta_2$}
    \KwRequire{Block size: $b$}
    \KwRequire{Hessian Power: $k$}
    Set: $m_0 = 0$, $v_0 =0$
    
    \For(\ \ \tcp*[h]{Training Iterations}){t $=1,2,\ldots $}{
    $\g_t \leftarrow$  current step gradient 
    
    $\mD_t \leftarrow$  current step estimated diagonal Hessian 
    
    Compute $\mD^{(s)}_t$ based on~\eref{eq:spatial_averaging}
    
    Update $\bar\mD_t$ based on~\eref{eq:hessian_momentum}
    
    Update $m_t,~v_t$ based on~\eref{eq:update_gh_adahess}
    
    $\theta_t = \theta_{t-1} - \eta m_t/v_t $
}
\end{algorithm}

\subsection{Hessian Momentum}
\label{sec:hessian_momentum}

We can easily apply momentum to Hessian diagonal since it is a vector instead of a 
quadratically large matrix. This enables us to adopt momentum for 
Hessian diagonal in \OURS. 
More specifically, let $\bar\mD_t$ denote the Hessian diagonal with momentum that is calculated
as:
\begin{equation}
\small
\label{eq:hessian_momentum}
\bar \mD_t = \sqrt{\frac{(1-\beta_2)\sum_{i=1}^t \beta_2^{t-i}\mD^{(s)}_i\mD^{(s)}_i}{1-\beta_2^t}}, 
\end{equation}
where $\mD^{(s)}$ is the spatially averaged Hessian diagonal (defined in~\eref{eq:spatial_averaging}),
and $0<\beta_2<1$ is the second moment hyperparameter. 
Note that this is exactly the same as the
momentum term in \adam~\cite{kingma2014adam} or RMSProp~\cite{tieleman2012lecture} except that we are using the spatial averaging Hessian diagonal instead of the gradient.

To illustrate the importance of Hessian momentum, we provide a simple example in 1D by 
considering $f (x) = x2 + 0.1x sin(20\pi x)$, as shown in~\fref{fig:2nd_momentum_illustration}. It can be clearly seen that the 
method without the second order momentum gets trapped at a local minima even with more than 1000
iterations (orange trajectory). On the contrary, the optimization converges within 7 iterations 
with Hessian momentum (blue trajectory). (While this example is over-simplified in certain ways,
we are using it here only to convey the importance of momentum.)

\subsection{AdaHessian}
\label{sec:adahessian_deeplearning}
To summarize, instead of only applying momentum for gradient, 
\OURS uses \textit{spatial averaging} and \textit{Hessian momentum} to smooth out local variations in Hessian diagonal.
More specifically, the first and second order moments ($m_t$ and $v_t$) for \OURS are computed as follows:
\begin{equation}
\small
\label{eq:update_gh_adahess}
\begin{split}
    m_t &= \frac{(1-\beta_1)\sum_{i=1}^t\beta_1^{t-i}\g_i }{1-\beta_1^t},\\
    v_t &= (\bar\mD_t)^k= \left(\sqrt{\frac{(1-\beta_2)\sum_{i=1}^t \beta_2^{t-i}\mD^{(s)}_i\mD^{(s)}_i}{1-\beta_2^t}}\right)^k, 
\end{split}
\end{equation}
where $0<\beta_1,~\beta_2<1$ are the first and second moment hyperparameters
that are also used in \adam.
Note that \adam uses the same
formulation except that the spatial averaging Hessian diagonal $\mD_i^{(s)}$ is replaced
with gradient.

The main overhead of \OURS is the Hutchinson's method to approximate Hessian diagonal, $\mD$.
We use one Hutchinson step per iteration to approximate the Hessian diagonal
(i.e., one random Rademacher vector $z$ in~\eref{eq:diag_estimation}).
The cost of this estimation is one Hessian matvec (to compute $Hz$),
which is equivalent to one gradient backpropagation~\cite{yao2018hessian,yao2019pyhessian}.

Also note that it is possible to get a more accurate approximation
to Hessian diagonal by using more Hutchinson steps per iteration. 
However, we found that one
step per iteration performs well in practice since the multiple calculations could be performed as Hessian momentum (\sref{sec:hessian_momentum}). 
In fact, as we discuss in~\sref{sec:result_delay_hessian_update}, it is possible to skip
the Hutchinson calculation for few iterations to reduce further its computational overhead, without
significant impact on final accuracy.

\section{Results}
\label{sec:results}
\subsection{Experiment Setup}
One of the problems with several formerly proposed optimization methods is that
the methods were originally tested with very simple models on very few tasks.
When those methods were later tested by the community on more complex models the results were
often worse than popular optimization methods. To avoid such a scenario, we extensively test
\OURS on a wide range of learning tasks, including image
classification, neural machine translation (NMT), language modeling (LM), and recommendation system (\recomsys). 
We compare the \OURS performance with
\sgd, \adam, \adamw~\cite{loshchilov2017decoupled}, and \adagrad.
Moreover, to enable a fair comparison we will use the same $\beta_1$ and $\beta_2$ parameters in \OURS as in \adam/\adamw
for each task, even though those default values may favor \adam (or \adamw) and disfavor \OURS. 
Furthermore, we will use the exact same weight decay and learning rate schedule in \OURS as that used by 
other optimizers. Below we briefly explain each of the learning tasks tested.

\begin{table}[t]
\caption{\footnotesize 
Results of ResNet20/32 on Cifar10 (left two columns) and ResNet18 on ImageNet (last column). 
On Cifar10:
\adam performs consistently worse than \sgd; 
\adamw has slightly worse performance than \sgd; and 
\OURS outperforms \adamw and even gets accuracy comparable to \sgd.
On ImageNet: 
\OURS has significantly better accuracy than \adam (5.53\%), \adamw (2.67\%), and has similar performance to \sgd.
}
\label{tab:cifar10_imagenet_result}
\centering
\begin{adjustbox}{width=1\linewidth} 
\centerline{
\begin{tabular}{lcccccc}
\toprule
\multirow{2}{*}{\textbf{Dataset}} & \multicolumn{2}{c}{\textbf{Cifar10}} & \textbf{ImageNet}      \\ 
 & \texttt{ResNet20} & \texttt{ResNet32} &\texttt{ResNet18}\\
\midrule
\sgd~\cite{ruder2016overview}            & 92.08 $\pm$ 0.08 & \bf{93.14} $\pm$ \bf{0.10}  & 70.03\\
\adam~\cite{kingma2014adam}              & 90.33 $\pm$ 0.13 & 91.63 $\pm$ 0.10       & 64.53\\
\adamw~\cite{loshchilov2017decoupled}    & 91.97 $\pm$ 0.15 & 92.72 $\pm$ 0.20       & 67.41\\
\midrule
\hc\OURS                                   & \bf{92.13} $\pm$ \bf{0.18} & 93.08 $\pm$ 0.10  & \bf{70.08}\\
\bottomrule
\end{tabular} 
}
\end{adjustbox}
\end{table}

\paragraph{Image Classification} 
We experiment on both Cifar10 (using ResNet20/32) and ImageNet (using ResNet18) datasets. 
Cifar10 consists of 50k training images and 10k testing images.
ImageNet has 1.2M training images and 50k validation images. 
We follow the settings described in~\cite{he2016deep} for training.
We run each experiment 5 times on Cifar10 and report the mean and standard deviation of the results. 

\paragraph{Neural Machine Translation (NMT)}
We use IWSLT14 German-to-English (De-En) and WMT14 English-to-German (En-De) datasets. 
Transformer \texttt{base} architecture is used for WMT14 (4.5M sentence pairs), and \texttt{small} architecture is used for IWSLT14 (0.16M sentence pairs). 
We follow the settings reported in~\cite{ott2018scaling} and use  pre-normalization described in~\cite{wang2019learning}. 
The length penalty is set to 0.6/1.0 and the beam size is set to 4/5 for WMT/IWSLT~\cite{ott2019fairseq}. 
We report the average results of the last 10/5 checkpoints respectively.
For NMT, BLEU score is used~\cite{papineni2002bleu}. 
In particular, we report tokenized case-sensitive BLEU on WMT14 En-De and case-insensitive BLEU IWSLT14 De-En.
Furthermore, we use \adamw for this task instead of \adam since the former is the standard optimizer (\adam consistently
scores lower).

\paragraph{Language Modeling} 
We use PTB~\cite{mikolov2011empirical} and Wikitext-103~\cite{merity2016pointer} datasets,
which contain 0.93M and 100M tokens, respectively. 
Following~\cite{ma2019tensorized}, a three-layer tensorized transformer core-1 for PTB and a six-layer tensorized transformer core-1 for Wikitext-103 are used in the experiments.
We apply the multi-linear attention mechanism with masking and report the perplexity~(PPL) on the test set with the best validation~model.

\begin{table}[!tbp]
\caption{\footnotesize 
NMT performance (BLEU) on IWSLT14 De-En and WMT14 En-De testsets (higher is better). 
Unlike in~\tref{tab:cifar10_imagenet_result}, \sgd has significantly worse
results than \adamw.
Note that \OURS outperforms the default and heavily tuned optimizer \adamw by 0.13 and 0.33 on IWSLT14 and WMT14, which is significant for this task. 
}
\label{tab:translation}
\centering
\begin{adjustbox}{width=0.8\linewidth} 
\begin{tabular}{lcccc}
\toprule
\multirow{2}{*}{\textbf{Model}} & \textbf{IWSLT14} & \multicolumn{1}{c}{\textbf{WMT14}}      \\ 
 & \texttt{small} & \texttt{base} \\
\midrule

\sgd                                        & 28.57 $\pm$ .15      & 26.04\\
\adamw~\cite{loshchilov2017decoupled}        & 35.66 $\pm$ .11     & 28.19 \\
\midrule
\hc\OURS                                       & \bf{35.79} $\pm$ .06    & \bf{28.52} \\
\bottomrule
\end{tabular}
\end{adjustbox}
\end{table}

\paragraph{Natural Language Understanding}
We use the GLUE task~\cite{wang2018glue} to evaluate the fine-tuning performance of SqueezeBERT~\cite{iandola2020squeezebert}. 
More specifically, we use 8 different tasks in GLUE and report and final average performance on the validation dataset.

\paragraph{Recommendation System} 
The Criteo Ad Kaggle dataset contains approximately 45 million samples over 7 days. 
We follow the standard setting and use the first 6 days as the training set and the last day as the test set. 
Furthermore, we use DLRM, a novel recommendation model that has been recently released by Facebook~\cite{naumov2019deep}.
The testing metric for Recommendation Systems is Click Through Rate (CTR), measured on training and test sets. 

We refer the interested reader to Appendix~\ref{sec:app_experimental_setup} for more detailed experimental settings.
Next we report the experimental results on each of these tasks.

\subsection{Image Classification}
\label{sec:result_image_classification}
The results on Cifar10 are shown in~\tref{tab:cifar10_imagenet_result}.
First, note the significantly worse performance of 
\adam, as compared to \sgd even on this simple image classification dataset. 
Particularly, \adam has 1.75\%/1.51\% lower accuracy for ResNet20/32 than \sgd. 
\adamw achieves better results than \adam, but its performance is still slightly worse than \sgd. 
However, \OURS achieves significantly better results as compared to \adam (1.80\%/1.45\% for ResNet20/32), 
even though we use the same $\beta_1$ and $\beta_2$ parameters in \OURS as in \adam.
That is, we did not tune these two hyperparameters, even though tuning them could potentially lead to even better performance.\footnote{In fact, in~\tref{tab:hessian_delay_update} we achieve 92.40
for ResNet20 which is higher than what we report in~\tref{tab:cifar10_imagenet_result}.
This is to emphasize that we only tuned learning rate in~\tref{tab:cifar10_imagenet_result}.
Still \OURS achieves
significantly better results than \adam.} 
Compared with SGD, \OURS achieves comparable accuracy for both ResNet20 (0.05\% higher) and ResNet32 (0.06\% lower).
The training and testing curves of different optimizers for ResNet20/32 on Cifar10 are shown in~\fref{fig:cifar10_training_testing_curve}.

Next, we use the best learning rate obtained by training ResNet20/32 on Cifar10 to optimize ResNet18 on ImageNet for all four optimizers.
We try two different learning rate schedules for all four optimizers, and we use the one with the better result. 
The two learning rate schedules are quite standard, i.e., the step decay schedule and the plateau based schedule~\cite{NEURIPS2019_9015}. 
The final result is reported in~\tref{tab:cifar10_imagenet_result}.
Again note that the final performances of \adam and \adamw are much worse than that of \sgd and \OURS. 
We plot the training and testing curve in~\fref{fig:imagenet_training_testing_curve}. 

It is worthwhile to note that our learning rate tuning is performed at an academic scale, but
\OURS still significantly
exceeds other adaptive methods and reaches the same performance level as \sgd which has been
tuned at the industrial scale.

\begin{table}[!tbp]
\caption{\footnotesize 
LM performance (PPL) on PTB and Wikitext-103 test datasets (lower is better). 
The PPL of \OURS is 2.7 and 1.0 lower than that of \adamw. 
}
\label{tab:language_modeling}
\centering
\begin{adjustbox}{width=0.9\linewidth} 
\begin{tabular}{lcccc}
\toprule
\multirow{2}{*}{\textbf{Model}} & \textbf{PTB} & \multicolumn{1}{c}{\textbf{Wikitext-103}}      \\
                                & \texttt{Three-Layer}  & \texttt{Six-Layer}              \\ 
\midrule
\sgd                                         & 59.9 $\pm$ 3.0     & 78.5 \\
\adamw~\cite{loshchilov2017decoupled}        & 54.2 $\pm$ 1.6     & 20.9 \\
\midrule
\hc\OURS                                       & \bf{51.5} $\pm$ 1.2 & \bf{19.9} \\
\bottomrule
\end{tabular}  
\end{adjustbox}
\end{table}

\begin{table*}[!t]
\caption{\footnotesize 
Comparison of \adamw and \OURS for SqueezeBERT on the development set of the GLUE benchmark. 
As can be seen, the average performance of \OURS is 0.41 higher as compared to \adamw. 
The result of $\text{\adamw}^+$ is directly from~\cite{iandola2020squeezebert} and the result of $\text{\adamw}^*$ is reproduced by us.
}
\label{tab:squeezebert_result}
\centering
\centerline{
\begin{tabular}{lccccccccc}
\toprule
\ha        & \textbf{RTE} & \textbf{MPRC} & \textbf{STS-B} & \textbf{SST-2} & \textbf{QNLI} & \textbf{QQP} & \textbf{MNLI-m} & \textbf{MNLI-mm} & Avg.
 \\ 
\midrule
\ha $\text{\adamw}^+$~\cite{iandola2020squeezebert}  & 71.8 & 89.8 & 89.4 & \bf{92.0} & \bf{90.5} & 89.4 & 82.9 & 82.3 & 86.01 \\
\midrule
\ha $\text{\adamw}^*$  & 79.06 & 90.69 & 90.00 & 91.28 & 90.30 & \bf{89.49} & 82.61 & 81.84 & 86.91 \\
\hc \OURS   & \bf{80.14} & \bf{91.94} & \bf{90.59} & 91.17 & 89.97 & 89.33 & \bf{82.78} & \bf{82.62} & \bf{87.32} \\
\bottomrule
\end{tabular} 
}
\end{table*}

\subsection{Neural Machine Translation}
\label{sec:result_machine_translation}

We use BLEU~\cite{papineni2002bleu} as the evaluation metric for NMT.
Following standard practice, we measure tokenized case-sensitive BLEU and case-insensitive BLEU for WMT14 En-De and IWSLT14 De-En, respectively. 
For a fair comparison, we do not include other external datasets.

The NMT results are shown in~\tref{tab:translation}. 
The first interesting observation is that here \sgd performs much worse than \adamw (which is opposite to its behaviour for image classification problems where \sgd has superior performance; see Appendix~\ref{sec:result_image_classification}). 
As pointed out in the introduction, even the choice of the optimizer has
become another hyperparameter. In particular, note that
the BLEU scores of \sgd are 7.09 and 2.15 lower than \adamw on IWSLT14 and WMT14, which is quite significant. 
Similar observations about \sgd were also reported in~\cite{zhang2019adam}.

Despite this, \OURS achieves state-of-the-art performance for NMT with transformers. In particular, 
\OURS outperforms \adamw by 0.13 BLEU score on IWSLT14.
Furthermore, the accuracy of \OURS on WMT14 is 28.52, which is 0.33 higher than that of \adamw. 
We also plot the training losses 
of \adamw and \OURS on IWSLT14/WMT14 in~\fref{fig:mt_training_curve}. 
As one can see, \OURS
consistently achieves lower training loss.
These improvements are quite significant for NMT, and importantly these are achieved even though \OURS
directly uses the same $\beta_1$ and $\beta_2$, as well as the same number of warmup iterations as in \adamw.

\subsection{Language Modeling}
\label{sec:language_modeling}

We report the language modeling results in~\tref{tab:language_modeling}, using the tensorized transformer proposed in~\cite{ma2019tensorized}. 
Similar to NMT, note that the perplexity (PPL) of \sgd is more than 57 points worse than \adamw on Wikitext-103. That is, similar to the NMT task, \sgd performs worse than \adamw.
However, \OURS achieves more than 1.8/1.0 better PPL  than that of \adamw on PTB/Wikitext-103, respectively. 

We also show the detailed training loss curves in~\fref{fig:ptb_training_curve}. 
\OURS achieves consistently lower loss values than \adamw throughout the training process on both PTB and Wikitext-103. 
Similar to NMT, the $\beta_1/\beta_2$ as well as the warmup phase of \OURS are kept the same as~\adamw.

\subsection{Natural Language Understanding}
\label{sec:natrual_language_understading}
We report the NLU results in~\tref{tab:squeezebert_result}, using the SqueezeBERT model~\cite{iandola2016squeezenet} tested on GLUE datasets~\cite{wang2018glue}. 
As can be seen, \OURS has better performance than \adamw on 5 out of 8 tasks. 
Particularly, on RTE and MPRC, \OURS achieves more than 1 point as compared to \adamw. 
On average, \OURS outperforms \adamw by 0.41 points. 
Note that similar to NMT and LM, except learning rate and block size, \OURS directly uses the same hyperparameters as \adamw. Interestingly, note that these results are better than
those reported in SqueezeBERT~\cite{iandola2020squeezebert}, even though
we only change the optimizer to \OURS instead of~\adamw.

\begin{figure}[!htbp]
\begin{center}
  \includegraphics[width=.98\linewidth]{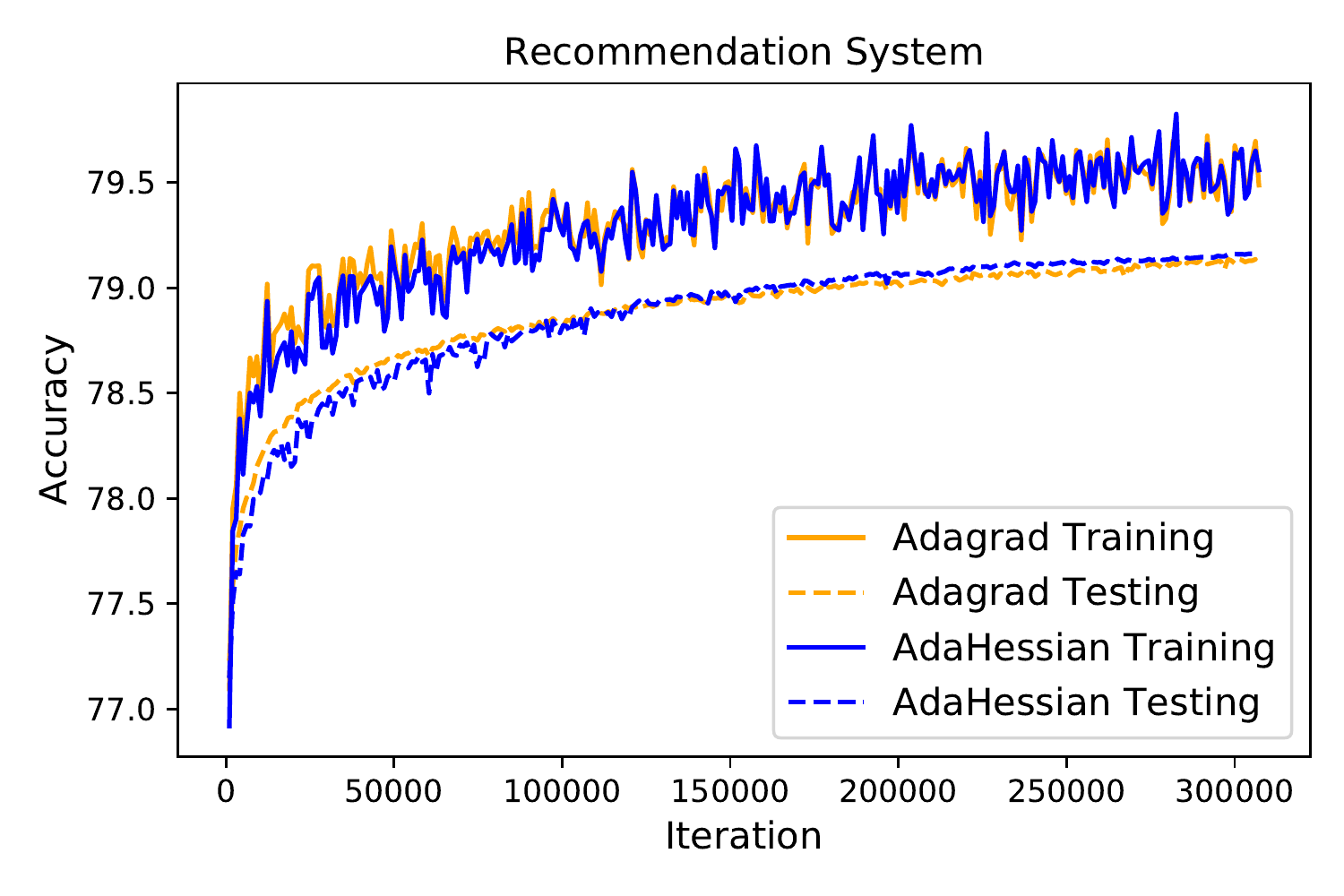}
\end{center}
 \caption{\footnotesize 
 Training and Testing Accuracy curves of \adagrad and \OURS on Criteo Ad Kaggle dataset. 
  As can be seen, the test accuracy of \OURS is better (0.032\%) than that of \adagrad.
  This is quite significant for this task.
 }
\label{fig:rs_training_test_curve}
\end{figure}

\subsection{Recommendation System}
\label{sec:recommendation_system}
We solely focus on modern recommendation systems, and in particular on the DLRM model
widely adopted in industry~\cite{naumov2019deep}.
These systems include a large embedding layer followed by a series of dense FC layers.
In training, a sparse set of rows of the embedding layer is used and only those rows are updated.
These rows do change from one iteration to the next. 
For such a sparse setting, we use \adagrad to update the embedding table, and we use \OURS to update the rest of the FC network in the experiments. 
(Pytorch currently does not support second order backpropagation for the sparse gradient to the embedding.) 
\OURS uses the same hyperparameters for updating the embedding table as in the \adagrad experiment without tuning. 
The training and testing accuracy curves are reported in~\fref{fig:rs_training_test_curve}. 
The testing accuracy of \OURS is 79.167\%, which is 0.032\% higher than \adagrad.
It should be noted that this is a quite significant accuracy increase for Recommendation Systems~\cite{wang2017deep}.


\begin{table*}[!t]
\caption{\footnotesize 
Robustness of \adamw and \OURS to the learning rate on IWSLT14. We scale the base learning rate used in~\sref{sec:result_machine_translation}. 
As can be seen, \OURS is much more robust to large learning rate variability as compared to \adamw. 
}
\label{tab:lr_stable_iwslt}
\centering
\small
\setlength\tabcolsep{3.pt}
\centerline{
\begin{tabular}{lcccccccccccc}
\toprule
LR Scaling               & \textbf{0.5} & \textbf{1} & \textbf{2} & \textbf{3} & \textbf{4} & \textbf{5} & \textbf{6} & \textbf{10} \\ 
\midrule
\ha\adamw                & \bf{35.42 $\pm$ .09} & 35.66 $\pm$ .11 & \bf{35.37 $\pm$ .07} & \bf{35.18 $\pm$ .07} & \bf{34.79 $\pm$ .15} & 14.41 $\pm$ 13.25 & 0.41 $\pm$ .32 & Diverge \\
\hc\OURS                 & 35.33 $\pm$ .10 & \bf{35.79 $\pm$ .06} & 35.21 $\pm$ .14 & 34.74 $\pm$ .10 & 34.19 $\pm$ .06 & \bf{33.78 $\pm$ .14} &  \bf{32.70 $\pm$ .10} & \bf{32.48 $\pm$ .83}       \\
\bottomrule
\end{tabular} 
}
\end{table*}
\begin{table*}[!t]
\caption{\footnotesize 
Block Size effect of \OURS on IWSLT14. With various block sizes, the performance of \OURS is very stable and no worse than that of \adamw (35.66 $\pm$ .11). 
}
\label{tab:block_stable_iwslt}
\centering
\small
\setlength\tabcolsep{3.pt}
\centerline{
\begin{tabular}{lccccccccc}
\toprule
\ha Block Size                   & \textbf{1}  & \textbf{2} & \textbf{4} & \textbf{8} & \textbf{16} & \textbf{32} & \textbf{64} & \textbf{128} \\ 
\midrule
\hc \OURS                        & 35.67 $\pm$ .10 & 35.66 $\pm$ .07 & 35.78 $\pm$ .07 & 35.77 $\pm$ .08 & 35.67 $\pm$ .08 & \bf{35.79 $\pm$ .06} & 35.72 $\pm$ .06 & 35.67 $\pm$ .11          \\
\bottomrule
\end{tabular} 
}
\end{table*}

\section{Discussion}
As reported in the previous section, \OURS achieves state-of-the-art performance on a wide
range of tasks. 
Two important issues are the sensitivity of \OURS to the hyperparameters of learning rate and block size. 
This is discussed next.

\subsection{Learning Rate and Block Size Effects}
\label{sec:blocksize_lr_effect}
Here, we explore the effects of the learning rate and block size $b$ on \OURS.
We first start with the effect of learning rate, and test the performance of \OURS and \adamw with different learning rates.
The results are reported in~\tref{tab:lr_stable_iwslt} for IWSLT14 dataset, where we scale the
original learning rate with a constant factor, ranging from 0.5 to 20 (the original learning rate is the same as in~\sref{sec:result_machine_translation}).
It can be seen that \OURS is more robust to the large learning rates. 
Even with $10\times$ learning rate scaling, \OURS still achieves 32.48 BLEU score, while \adamw diverges even with $6\times$ learning rate scaling. 
This is a very desirable property of \OURS, as it results in reasonable performance for such
a wide range of learning rates.

We also test the effect of the spatial averaging block size (parameter $b$ in~\eref{eq:spatial_averaging}). As a reminder, this parameter is used for spatially averaging the Hessian diagonal as illustrated in~\fref{fig:block_hessian_vis_conv_multihead}. The sensitivity results are shown in~\tref{tab:block_stable_iwslt} where we vary
the block size from 1 to 128.
While the best performance is achieved for the block size of 32, the performance variation for other block sizes is rather small. 
Moreover, all the results are still no worse than the result with \adamw.

\begin{table*}[!t]
\caption{\footnotesize 
Comparison between \OURS theoretical and measured speed, as compared to \adam
and \sgd, tested on Cifar10. 
We also measured the speed up for different Hessian computation frequencies.
As one can see, \OURS is \emph{not} orders of magnitude slower than \sgd, despite the widely-held incorrect belief about the efficiency of Hessian based methods. 
Furthermore, by increasing the Hessian computation frequency,
the run time can improve from $3.23\times$ to $1.45\times$, as compared to \sgd for ResNet32.
The real measurement is performed on one RTX Titan GPU. 
}
\label{tab:hessian_delay_update}
\centering
\centerline{
\begin{tabular}{lcccccc}
\toprule
Hessian Computation Frequency   & \textbf{1}        & \textbf{2}        & \textbf{3}        & \textbf{4}        & \textbf{5}  \\ 
\midrule
\ha Theoretical Per-iteration Cost ($\times$\sgd)  & 2$\times$  & 1.5$\times$       &1.33$\times$       &1.25$\times$       &1.2$\times$\\
\midrule
ResNet20 (Cifar10)             & 92.13 $\pm$ .08   & 92.40 $\pm$ .04   & 92.06 $\pm$ .18   & 92.17 $\pm$ .21   & 92.16 $\pm$ .12\\
\hb Measured Per-iteration Cost ($\times$\sgd)     & 2.42$\times$ & 1.71$\times$   & 1.47$\times$ & 1.36$\times$ & 1.28$\times$\\
\hc Measured Per-iteration Cost ($\times$\adam)     & 2.27$\times$ & 1.64$\times$   & 1.42$\times$ & 1.32$\times$ & 1.25$\times$\\
\midrule
ResNet32 (Cifar10)             & 93.08 $\pm$ .10  & 92.91 $\pm$ .14   & 92.95 $\pm$ .17   & 92.93 $\pm$ .24   & 93.00 $\pm$ .10 \\
\hb Measured Per-iteration Cost ($\times$\sgd)     & 3.23$\times$ & 2.12$\times$ & 1.74$\times$ & 1.56$\times$ & 1.45$\times$\\
\hc Measured Per-iteration Cost ($\times$\adam)     & 2.91$\times$ & 1.96$\times$ & 1.64$\times$ & 1.48$\times$ & 1.38$\times$\\
\bottomrule
\end{tabular} 
}
\end{table*}

\subsection{\OURS Overhead}
\label{sec:result_delay_hessian_update}
Here, we discuss and measure the overhead of \OURS.
In terms of computational complexity, \OURS requires twice the flops as compared to \sgd. 
This $2\times$ overhead comes from the cost of computing the Hessian diagonal, when
one Hutchinson step is performed per optimization iteration.
Each Hutchinson step requires computing one Hessian matvec (the $\H z$ term in~\eref{eq:diag_estimation}).
This step requires one more gradient backpropagation, hence leading to twice the theoretical complexity. 

We have also measured the actual runtime of \OURS in PyTorch on a single RTX Titan GPU machine, as
reported in the second column of~\tref{tab:hessian_delay_update}.
For ResNet20, \OURS is $2.42\times$ slower than \sgd (and $2.27\times$ slower than \adam).
As one can see, \OURS is not orders of magnitude slower than first order methods.
The gap between the measured and theoretical speed is likely due to the fact that Pytorch~\cite{paszke2017automatic} (and other existing frameworks) are highly optimized for first order methods.
Even then, if one considers the fact that \sgd needs a lot of tuning, this overhead
may not be large.

It is also possible to reduce
the \OURS overhead.
One simple idea is to reduce the Hutchinson calculation frequency
from 1 Hessian matvec per iteration to every multiple iterations. For example, for a frequency
of 2, we perform the Hutchinson step at every other optimization iteration. This reduces the theoretical
computational cost to $1.5\times$ from $2\times$. One can also further reduce the
frequency to 5, for which this cost reduces to $1.2\times$. 

We studied how such reduced Hutchinson calculation frequency approach would impact
the performance. We report the results for training ResNet20/ResNet32 on the
Cifar10 in~\tref{tab:hessian_delay_update}, when
we vary the Hutchinson frequency from 1 to 5. As one can see, there
is a small performance variation, but the \OURS overhead significantly decreases as compared to
\sgd and \adam. 
\section{Conclusions}
\label{sec:conclusions}
In this work, we proposed \OURS, an adaptive Hessian based optimizer.
\OURS incorporates an approximate Hessian diagonal, with spatial averaging and momentum to precondition the gradient vector. 
This automatically rescales the gradient vector
resulting in better descent directions.
One of the key novelties in our approach is the incorporation
spatial averaging for Hessian diagonal along with
an exponential moving average in time.
These enable us to smooth noisy local Hessian information which could be highly misleading.

We extensively tested \OURS on various datasets and tasks, using state-of-the-art models.
These include IWSLT14 and WMT14 for
neural machine translation, PTB and Wikitext-103 for language modeling, GLUE for natural language understanding, Cifar10 and ImageNet for image classification (provided in Appendix~\ref{sec:result_image_classification}), and Criteo Ad Kaggle for recommendation system (provided in Appendix~\ref{sec:recommendation_system}). 
\OURS consistently achieves comparable or higher generalization performance as
compared to the highly tuned default optimizers used for these different tasks. 

Stepping back, it is important for every work to state its limitations (in general, but in particular in this area).
The current limitation of \OURS is that it is $2-3\times$ slower than first order methods such as \sgd and \adam.
We briefly explored how this overhead could be reduced, but more work is needed
in this area.
However, \OURS consistently achieves comparable or better accuracy. 
For example, for LM task, \OURS achieves up to 2.7 better PPL, as compared to \adamw, which is significant for this task.

Finally, from a higher-level perspective, we should note that there has been significant development within second order methods, both theory and practice, even though these methods were widely viewed as being inapplicable for machine learning even just a few years ago. 
Some examples include Hessian based model compression~\cite{lecun1990optimal,hassibi1993second,dong2019hawqv2,dong2019hawq}, adversarial attacks~\cite{yao2019trust}, and studies of the loss landscape topology for different
NN architectures~\cite{santurkar2018does,yao2019pyhessian}, to name just a few.
\OURS is an important step in this area, and we expect that it will enable still further progress.
We have open sourced \OURS and we hope that it would help this progress~\cite{adahessian}.

\section*{Acknowledgments}
This work was supported by a gracious fund from Amazon Machine Learning Research Award (MLRA).
The UC Berkeley team also acknowledges gracious support from Intel corporation, Intel VLAB team, Google Cloud, Google TFTC team, and Nvidia.
Amir Gholami was supported through funding from Samsung SAIT.
Michael Mahoney would like to acknowledge the DARPA, NSF, and ONR via its BRC on RandNLA for providing partial support of this work.  Our conclusions do not necessarily reflect the position or the policy of our sponsors, and no official endorsement should be inferred.

{
\fontsize{9.5pt}{10.5pt} \selectfont
\bibliography{ref.bib}

}
\clearpage
\onecolumn
\appendix

\counterwithin{figure}{section}
\counterwithin{table}{section}

\subsection{Descending Property of~\eref{eq:gradient_hessian_direction}}
\label{sec:proof_gfh_direction}
Assume that $f(w)$ is a strongly convex and strictly smooth
function in $\sR^d$, such that there exists positive constants
$\alpha$ and $\beta$ so that  $ \alpha I\leq \nabla^2 f(w) \leq \beta I$ for all $w$.
We can show that the update
formulation of~\eref{eq:gradient_hessian_direction} is a converging algorithm. 
Particularly, we can show that with the proper learning rate:
\begin{equation}\label{eq:linear_convergence}
\small
    f(w_{t+1}) - f(w_t) \leq -\frac{\alpha^k}{2\beta^{1+k}} \|\g_t\|^2.
\end{equation}
Note that when $k=0$ or $1$, the convergence rate is the same as gradient descent or Newton method\footnote{The convergence rate here denotes the global convergence rate of Newton's method.}~\cite{boyd2004convex}, respectively.
Our proof is similar to ~\cite{boyd2004convex} for Newton method. 

Let us define $\lambda(\wt) = (\gt^T\Ht^{-k}\gt)^{1/2}$. 
Since $f(w)$ is strongly convex, we have 
\begin{equation}
\begin{split}
    f(w_t-\eta\Deltawt) 
    &\leq f(\wt) - \eta\g_t^T\Deltawt + \frac{\eta^2\beta\|\Deltawt\|^2}{2} \\
    &\leq f(\wt) - \eta\lambda(\wt)^2 + \frac{\beta}{2\alpha^k}\eta^2\lambda(\wt)^2.
\end{split}
\end{equation}
The last inequality comes from the fact that
\begin{equation}
    \lambda(\wt)^2 = \Deltawt^T \H_t^k \Deltawt \geq \alpha^k \|\Deltawt\|^2. 
\end{equation}
Therefore, the step size $\hat\eta=\frac{\alpha^k}{\beta}$ will make $f$ decreases as follows,
\begin{equation}
    f(\wt-\hat\eta\Deltawt) \leq f(w_t) - \frac{1}{2}\hat\eta \lambda(\wt)^2.
\end{equation}
Since $\alpha\preceq \Ht\preceq \beta$, we have
\begin{equation}
    \lambda(\wt)^2 = \gt^T\Ht^{-k}\gt \geq \frac{1}{\beta^k} \|\g_t\|^2. 
\end{equation}
Therefore, 
\begin{equation*}
    f(\wt-\hat\eta\Deltawt) - f(\wt) \leq - \frac{1}{2\beta^k}\hat\eta\|\g_t\|^2 =  -\frac{\alpha^k}{2\beta^{1+k}}\|\g_t\|^2.
\end{equation*}

\subsection{Descending Property of~\eref{eq:gradient_diagonal_direction}}
\label{sec:proof_gdh_direction}
When $f(w)$ is a strongly convex and strictly smooth
function in $\sR^d$, such that there exists positive constants
$\alpha$ and $\beta$ so that  $ \alpha I\leq \nabla^2 f(w) \leq \beta I$ for all $w$,
we can prove that~\eref{eq:gradient_diagonal_direction} has the same convergence rate as~\eref{eq:gradient_hessian_direction}.

First of all, it is not hard to see the diagonal elements in $\mD$ are all positive since $f(w)$ is a strongly convex problem. 
That is, 
\begin{equation}
    \alpha \leq e_i^T \H e_i = e_i^T \mD e_i = D_{i,i},
\end{equation}
where $e_i$  is the vector whose coordinates are all zero, except the i-th one that equals 1. 
Similarly, we have  
\begin{equation}
    D_{i,i} = e_i^T \mD e_i = e_i^T \H e_i \leq \beta.
\end{equation}
Therefore, the diagonal elements in $\mD$ are in the range $[\alpha, \beta]$. 
Using the same proof as in Appendix~\ref{sec:proof_gfh_direction}, we will get the result. 

\subsection{Descending Property of~\eref{eq:spatial_averaging}}
\label{sec:proof_spatial_averaging}
When $f(w)$ is a strongly convex and strictly smooth
function in $\sR^d$, such that there exists positive constants
$\alpha$ and $\beta$ so that  $ \alpha I\leq \nabla^2 f(w) \leq \beta I$ for all $w$,
we can prove that~\eref{eq:spatial_averaging} has the same convergence rate as~\eref{eq:gradient_hessian_direction}.

As shown in Appendix~\ref{sec:proof_gdh_direction}, the diagonal elements in $\mD$ are in the range $[\alpha, \beta]$. 
Therefore, the average of a subset of those numbers is still in the range $[\alpha, \beta]$.
Using the same proof as in Appendix~\ref{sec:proof_gfh_direction}, we will get the result. 

\subsection{Experimental Setup}
\label{sec:app_experimental_setup}
Here, we provide more details on the experimental setup for the empirical evaluation.

\paragraph{Image Classification}
The training/test sets for Cifar10~\cite{krizhevsky2009learning} dataset contain 50k/10k images, respectively. 
The models used on Cifar10 are standard ResNet20/32. 
We train both models with 160 epochs and decay the learning rate by a factor of 10 at epoch 80 and 120. 
The batch size is set to be 256.
For \sgd/\adam/\adamw, the initial learning rates are tuned and set to be 0.1/0.001/0.01. 
For \OURS, we set the block size as 9, k to be 1, and learning rate as 0.15 for both ResNet20/32. 
For \adam/\adamw/\OURS, $\beta_1=0.9$ and $\beta_2=0.999$. 
We run each experiment 5 times on Cifar10 and report the mean and standard deviation of the results. 
The training/test sets for ImageNet dataset~\cite{deng2009imagenet} contain 1.2M/50k images, respectively. 
Our code is modified from the official PyTorch example\footnote{https://github.com/pytorch/examples/tree/master/imagenet}.
The batch size is set to be 256.
We train ResNet18 for 90 epochs. 
All the settings of different optimizers are the same as used in Cifar10 example.

\paragraph{Neural Machine Translation} 
The training/validation/test sets for the IWSLT14 dataset contain about 153K/7K/7K sentence pairs, respectively. 
We use a vocabulary of 10K tokens via a joint source and target byte pair encoding (BPE). 
For the WMT14 dataset, we follow the setting of~\cite{vaswani2017attention}, 
which contains 4.5M parallel sentence pairs for training. 
We use Newstest2014 as the test set, and Newstest2013 as the validation set. 
The 37K vocabulary for WMT14 is also via a joint source and target BPE factorization. 
We set dropout as $0.0$ for Transformer \texttt{base}/\texttt{small} model. 
For \adamw, we follow the optimizer setting and learning rate schedule in~\cite{wang2019learning}. 
For \OURS, we set the block size as 32 for IWSLT/WMT, k to be 1.0, and learning rate as 0.047/1.0 for IWSLT/WMT. 
For both \adamw/\OURS, we set $\beta_1=0.9$ and $\beta_2=0.98$.
We fix the label smoothing value as $\epsilon_\text{ls} = 0.1$ in all experiments.
We implement our code for MT based on \textit{fairseq-py}~\cite{ott2019fairseq}. 
We employ BLEU\footnote{https://github.com/moses-smt/mosesdecoder/blob/master/scripts/generic/multi-bleu.perl}~\cite{papineni2002bleu} as the evaluation metric for MT.
Following standard practice, we measure tokenized case-sensitive BLEU and case-insensitive BLEU for WMT14 En-De and IWSLT14 De-En, respectively. 
For a fair comparison, we do not include other external datasets. 
We train 130/55 epochs for WMT/IWSLT, respectively. 
We set the maximum token size to be $4096\times 8$ (eight gpus)/4096 (one gpu) for WMT/IWSLT. 
For inference, we average the last 10/5 checkpoints, and we set the length penalty as 0.6/1.0 and beam size as 4/5 for WMT/IWSLT, following~\cite{ott2019fairseq}. 
We run each experiment 5 times on IWSLT and report the mean and standard deviation of the results. 

\paragraph{Language Modeling}
PTB~\cite{mikolov2011empirical} has 0.93M training tokens, 0.073M validation tokens, and 0.082M test tokens. 
Wikitext-103~\cite{merity2016pointer} contains 0.27M unique words, and 100M training words from 28K articles, with an average length of 3.6K words per article. 
We use the same evaluation scheme following~\cite{dai2019transformer}. 
We use a three-layer tensorized transformer core-1 for PTB and a six-layer tensorized transformer core-1 for Wikitext-103, following~\cite{ma2019tensorized}.
We set the dropout rate as 0.3 in all the LM experiments. 
The model is trained for 30 epochs on both PTB and WikiText-103. 
For \adamw, we follow the learning rate setting in~\cite{ma2019tensorized}. 
For \OURS, we set the block size as 4 and k as 0.5 for PTB and Wikitext-103. We set the learning rate as 2.0/1.0 for PTB/Wikitext-103, respectively. 
We set the batch size to be 60/120 for PTB/Wikitext-103.
For \adamw/\OURS, $\beta_1=0.9$ and $\beta_2=0.999$.
We set the warmup steps to be 4000 and label smoothing to be $\epsilon_\text{ls} = 0.1$ in all LM experiments.
We run each experiment 5 times on PTB and report the mean and standard deviation of the results.

\paragraph{Natural Language Understanding}
The SqueezeBERT model is pre-trained as the authors suggested~\cite{iandola2020squeezebert}. 
Particularly, we pretrain SqueezeBERT from scratch using the LAMB~\cite{you2019large} optimizer
with a global batch size of 8192,
a learning rate of 2.5e-3, and a warmup proportion of 0.28. 
We pretrain for 56k steps with a maximum sequence length of 128 and then for
6k steps with a maximum sequence length of 512 followed~\cite{iandola2016squeezenet}.
Moreover, before directly using the pre-trained model on GLUE tasks, we apply transfer learning
from the MNLI GLUE task~\cite{wang2018glue} to other GLUE tasks~\cite{iandola2016squeezenet,phang2018sentence}. 
We refer readers to~\cite{iandola2016squeezenet} for more detailed instruction. 

For finetuning, we set the batch size as 16, $\beta_1=0.9$, and $\beta_2=0.999$ for both \adamw/\OURS as suggested in~\cite{iandola2016squeezenet}. 
For \OURS, we set the block size $b$ as 4 and k as 1 for all tasks. 
As in~\cite{iandola2016squeezenet} we perform hyperparameter tuning on the learning rate and dropout rate. 

\paragraph{Recommendation System}
The Criteo Ad Kaggle dataset consists of click logs for ad CTR prediction for 7 days. 
Each data set contains 13 continuous and 26 categorical features.
The dataset contains about 45 million samples over 7 days. 
In experiments, we follow the setting from~\cite{naumov2019deep}. 
Our code is also modified from~\cite{naumov2019deep}\footnote{https://github.com/facebookresearch/dlrm}. 
The testing metric for Recommendation Systems is Click Through Rate (CTR), measured on training and test sets. 
For \adagrad, the learning rate is set to be 0.01. 
For \OURS, we set the block size as 1, k as 0.5, learning rate as 0.043, $\beta_1=0.9$, and $\beta_2=0.98$.
We set the batch size to be 128, following~\cite{naumov2019deep}.

\paragraph{Delayed Hessian Update}
For ResNets on Cifar10, we use 5 epochs for warmup. 
In particular, within 5 epochs, the Hessian diagonal is still computed for every iteration. 
After that, the Hessian diagonal computation frequency is set to be between 1 to 5 iterations.

\subsection{Additional Results}
\begin{figure}[t]
\centering
  \includegraphics[width=.49\linewidth]{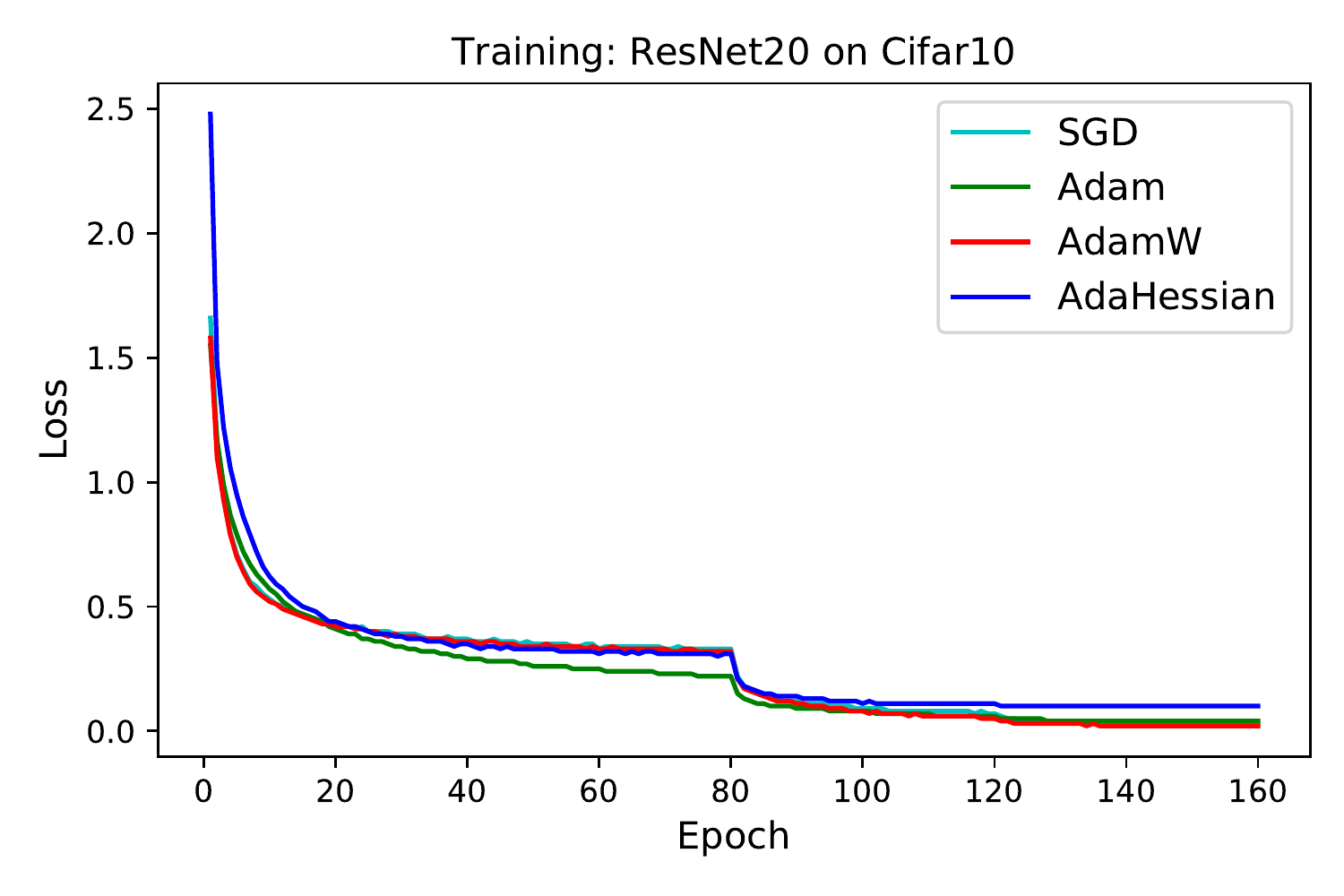}
  \includegraphics[width=.49\linewidth]{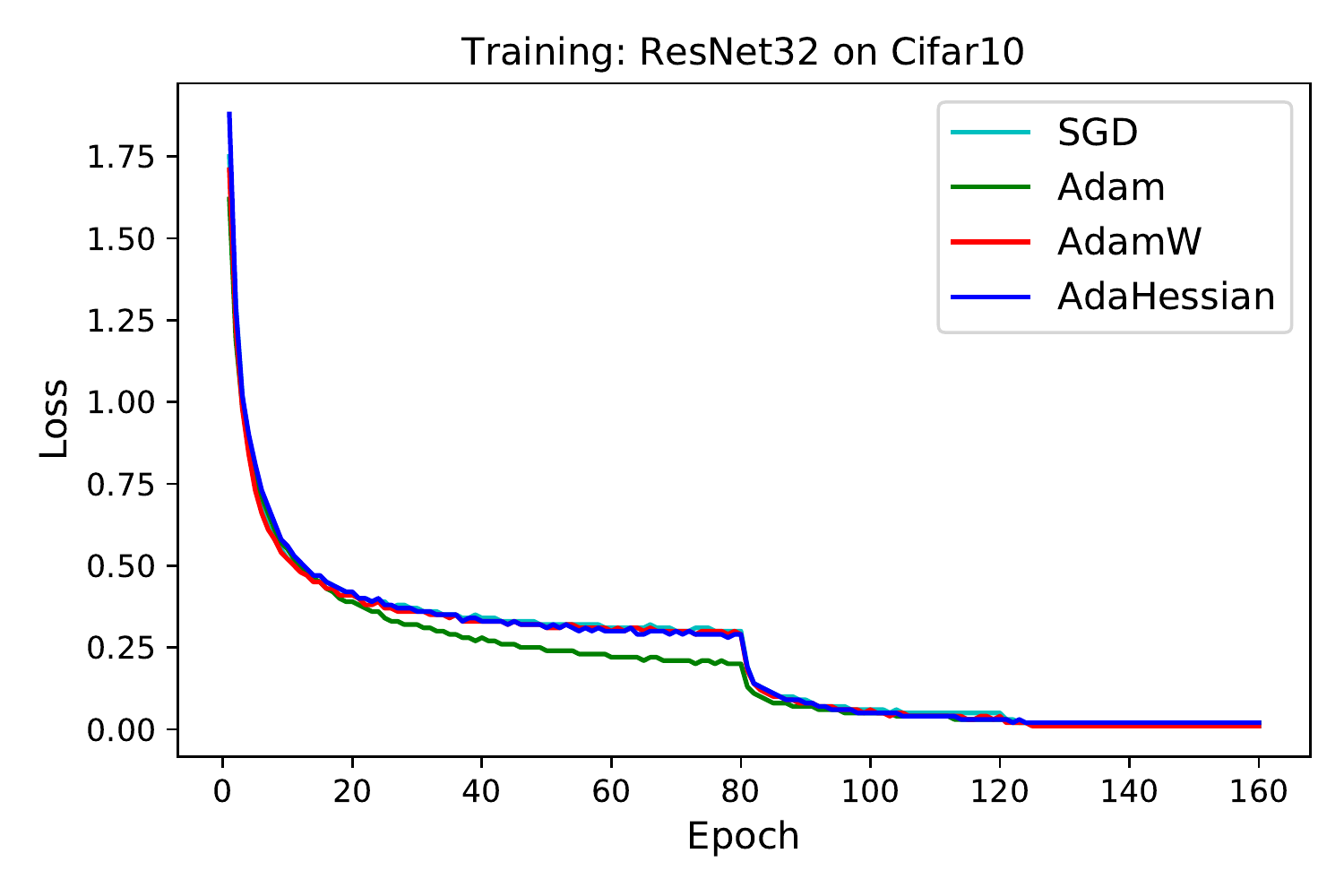}\\
  \includegraphics[width=.49\linewidth]{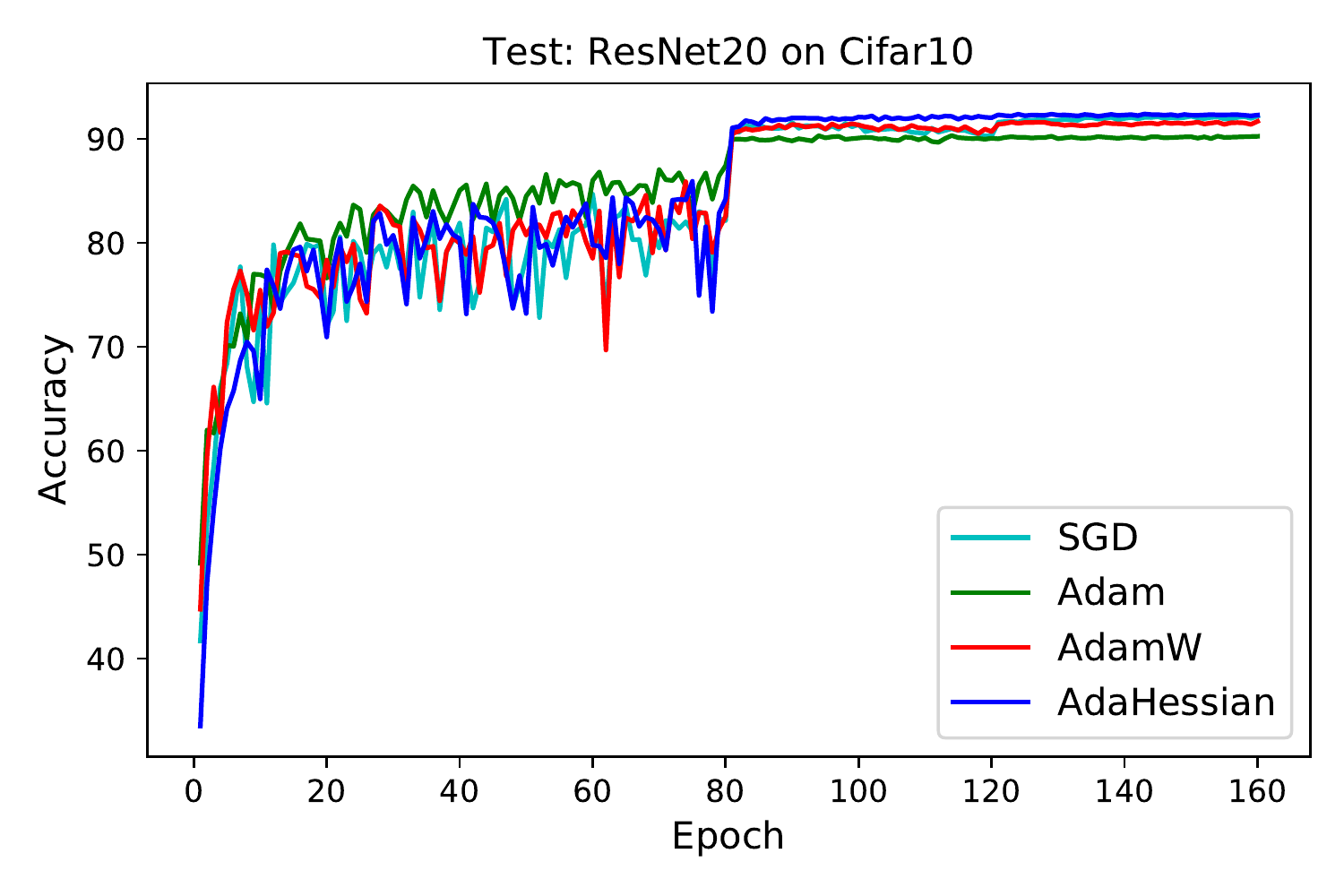}
  \includegraphics[width=.49\linewidth]{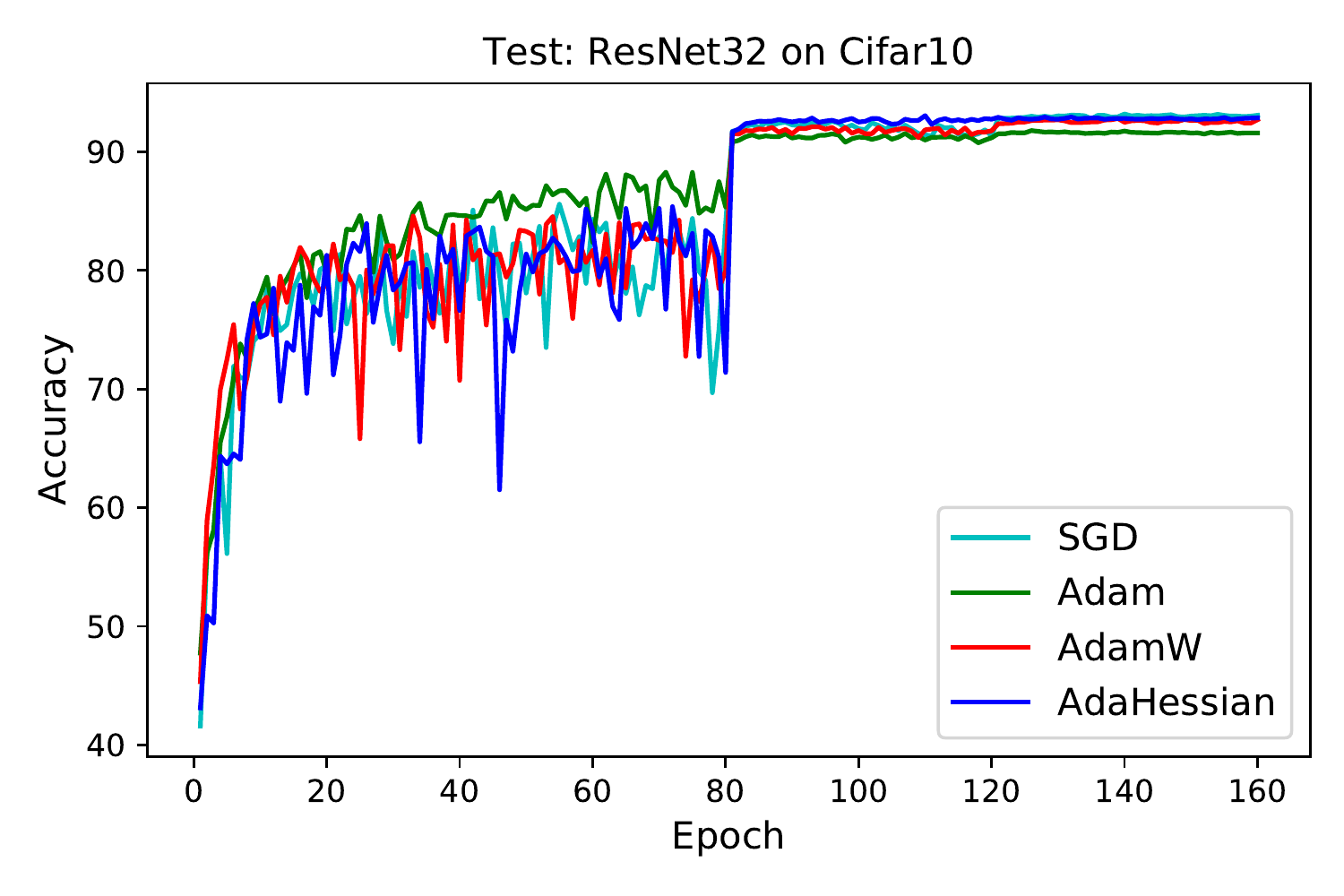}
 \caption{\footnotesize 
 Training and testing loss curves of \sgd, \adam, \adamw, \OURS for ResNet20/32 on Cifar10.
 \sgd and \OURS consistently achieve better accuracy as compared to \adam and \adamw. The 
 final accuracy results are reported in~\tref{tab:cifar10_imagenet_result}.
 }
\label{fig:cifar10_training_testing_curve}
\end{figure}

\begin{figure}[!ht]
\begin{center}
\includegraphics[width=.49\linewidth]{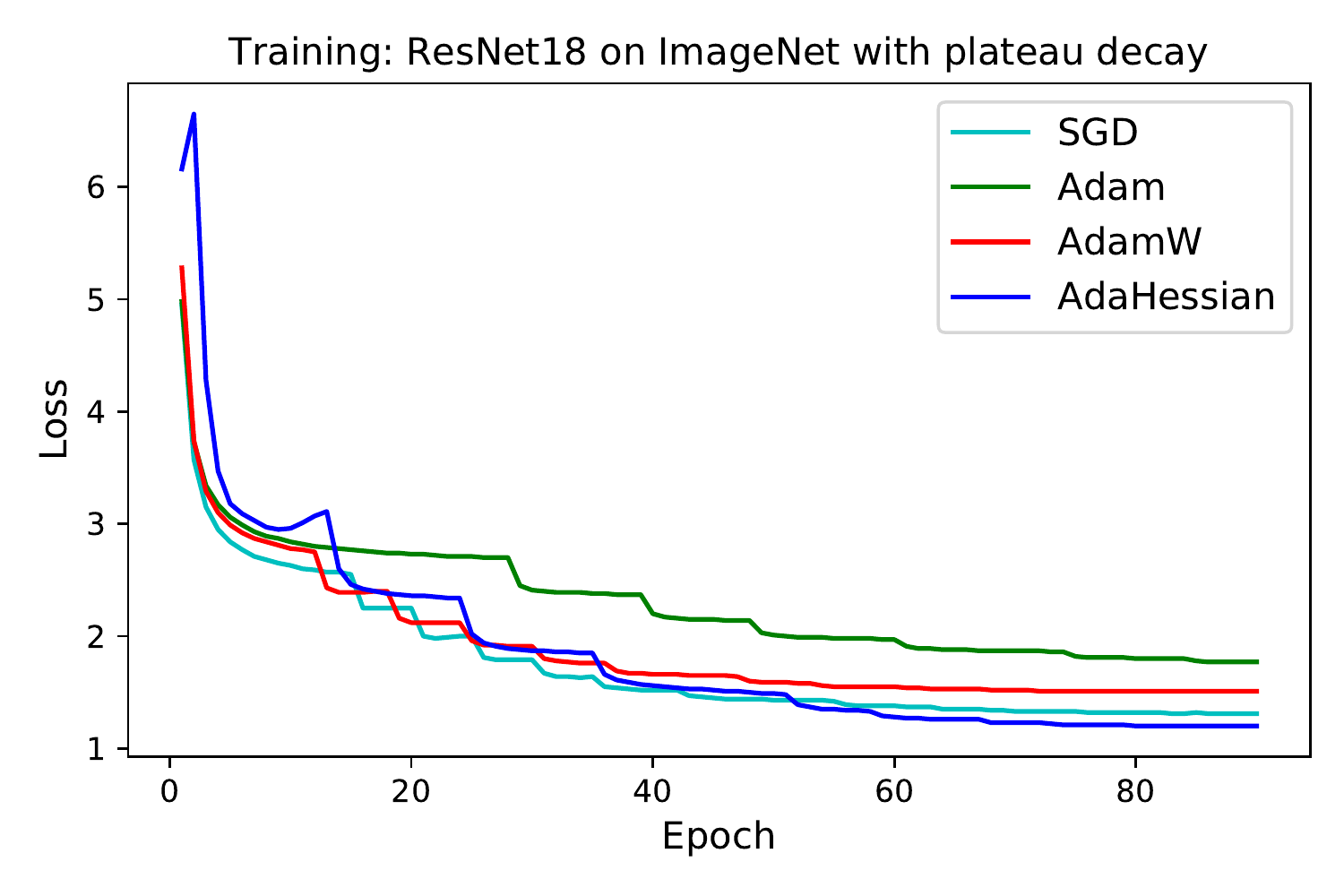}
  \includegraphics[width=.49\linewidth]{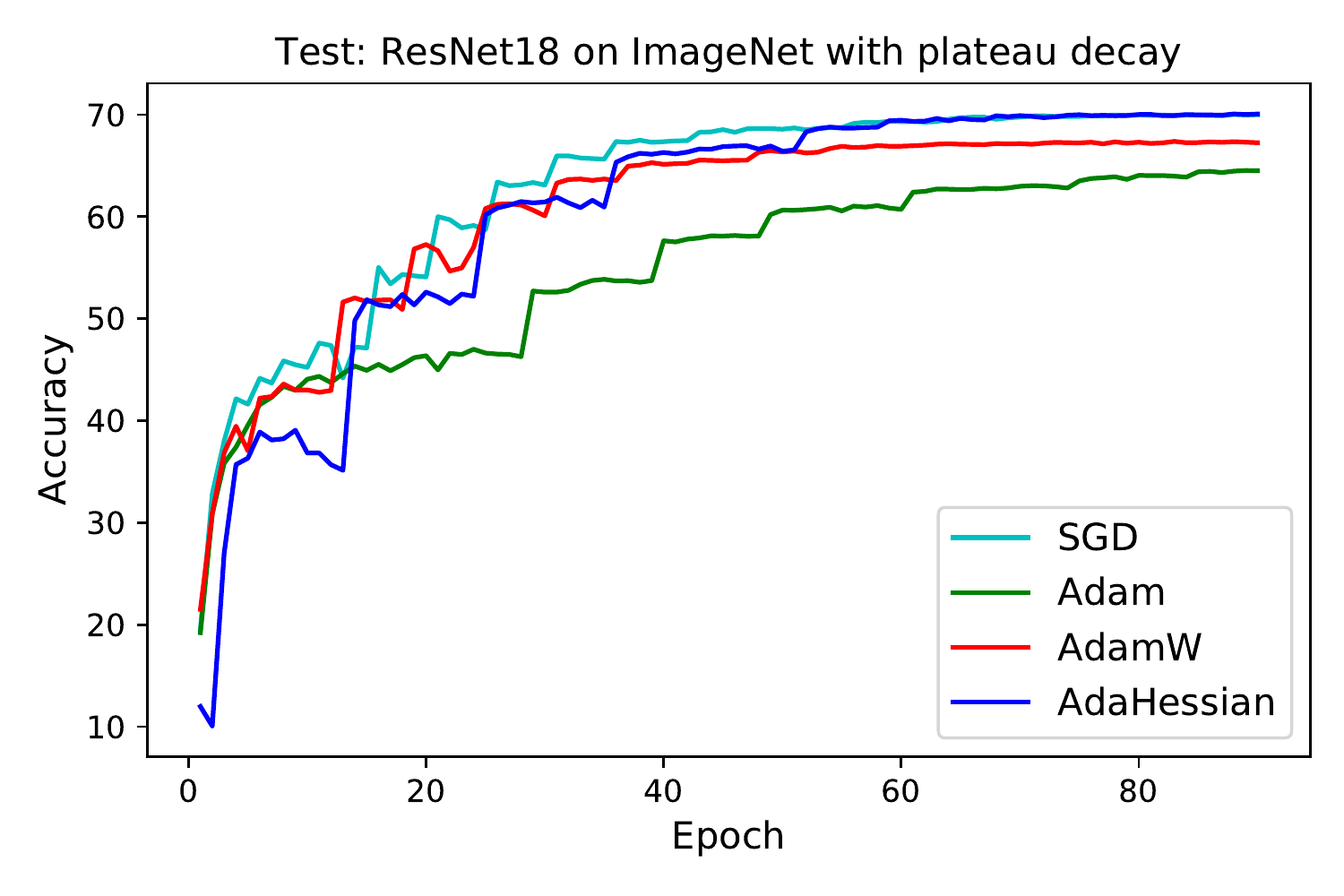}
\end{center}
 \caption{\footnotesize 
 Training/Test loss curve of \sgd, \adam, \adamw, \OURS for ResNet18 on ImageNet.
 \sgd and \OURS consistently achieve better accuracy as compared to \adam and \adamw.
 The 
 final accuracy results are reported in~\tref{tab:cifar10_imagenet_result}.
 }
\label{fig:imagenet_training_testing_curve}
\end{figure}

In this section, we present additional empirical results that were discussed in~\sref{sec:results}.
See~\fref{fig:cifar10_training_testing_curve} and~\ref{fig:imagenet_training_testing_curve}.

\begin{figure*}[!t]
\begin{center}
  \includegraphics[width=.45\linewidth]{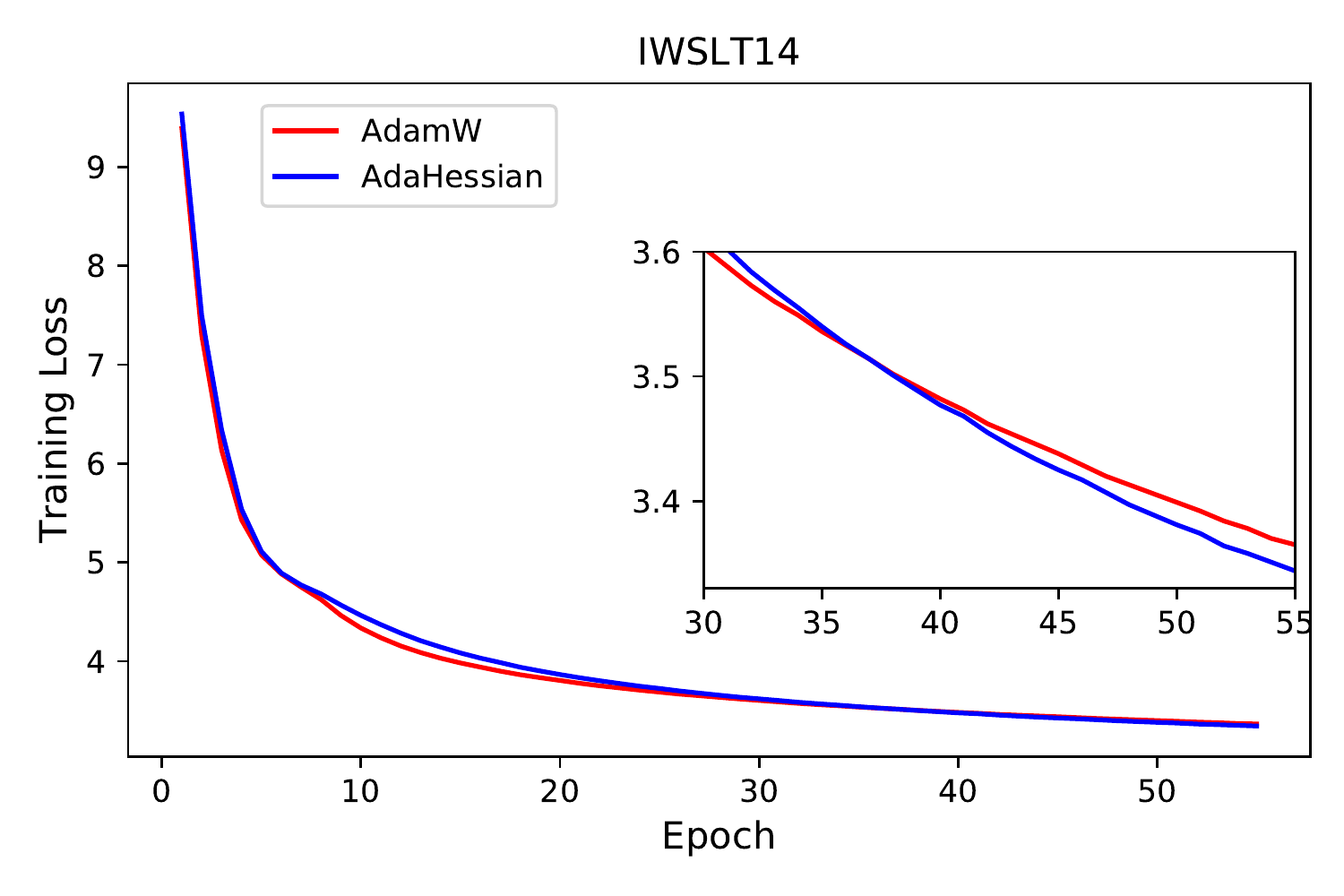}
  \includegraphics[width=.45\linewidth]{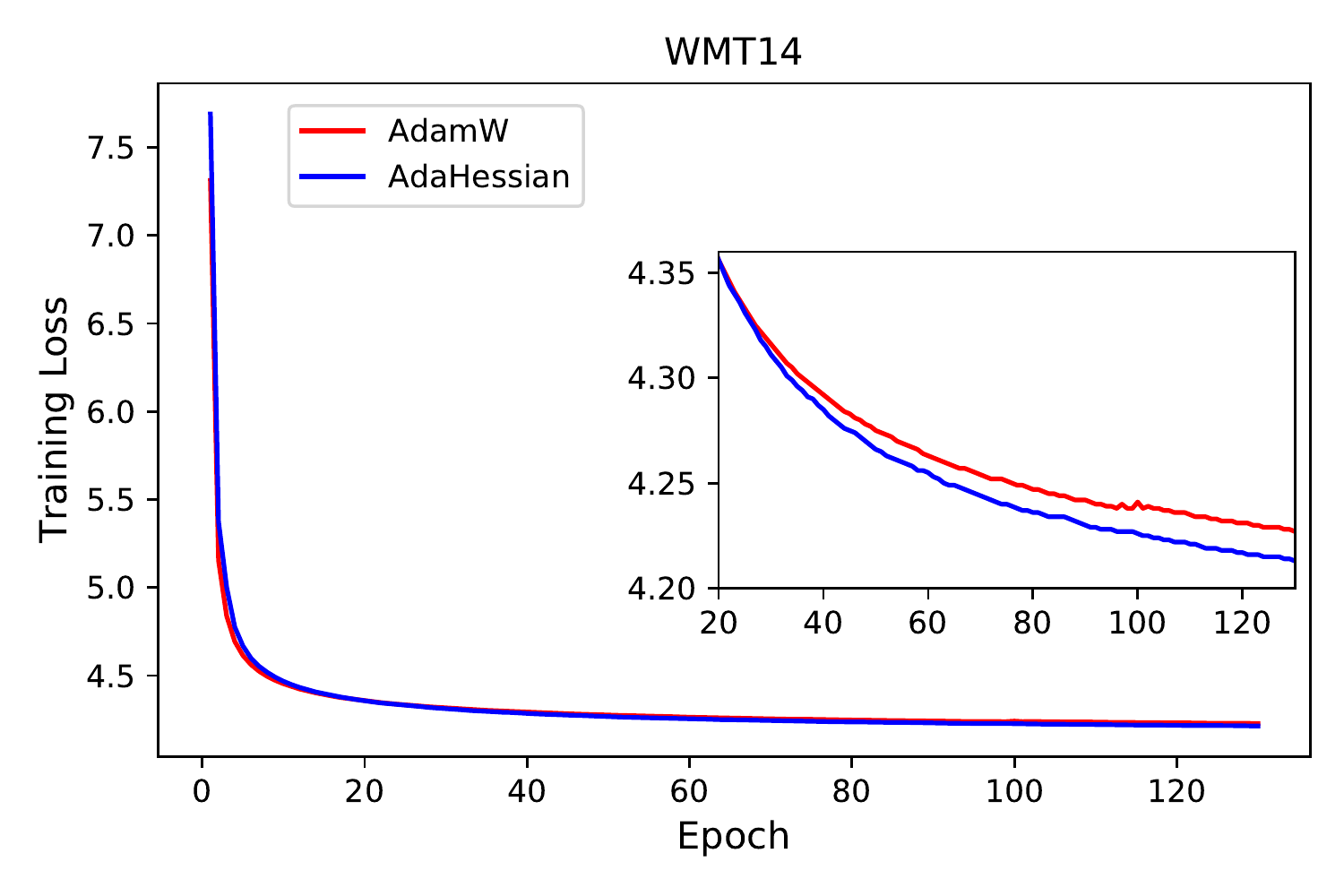}
\end{center}
 \caption{\footnotesize 
 Training loss curves of \adamw and \OURS for Transformer on IWSLT14 and WMT14. 
 The training loss of \OURS is lower than that of \adamw on both IWSLT14 and WMT14.
 Testing results are reported in~\tref{tab:translation}.
 }
\label{fig:mt_training_curve}
\end{figure*}

\begin{figure*}[t]
\begin{center}
  \includegraphics[width=.45\linewidth]{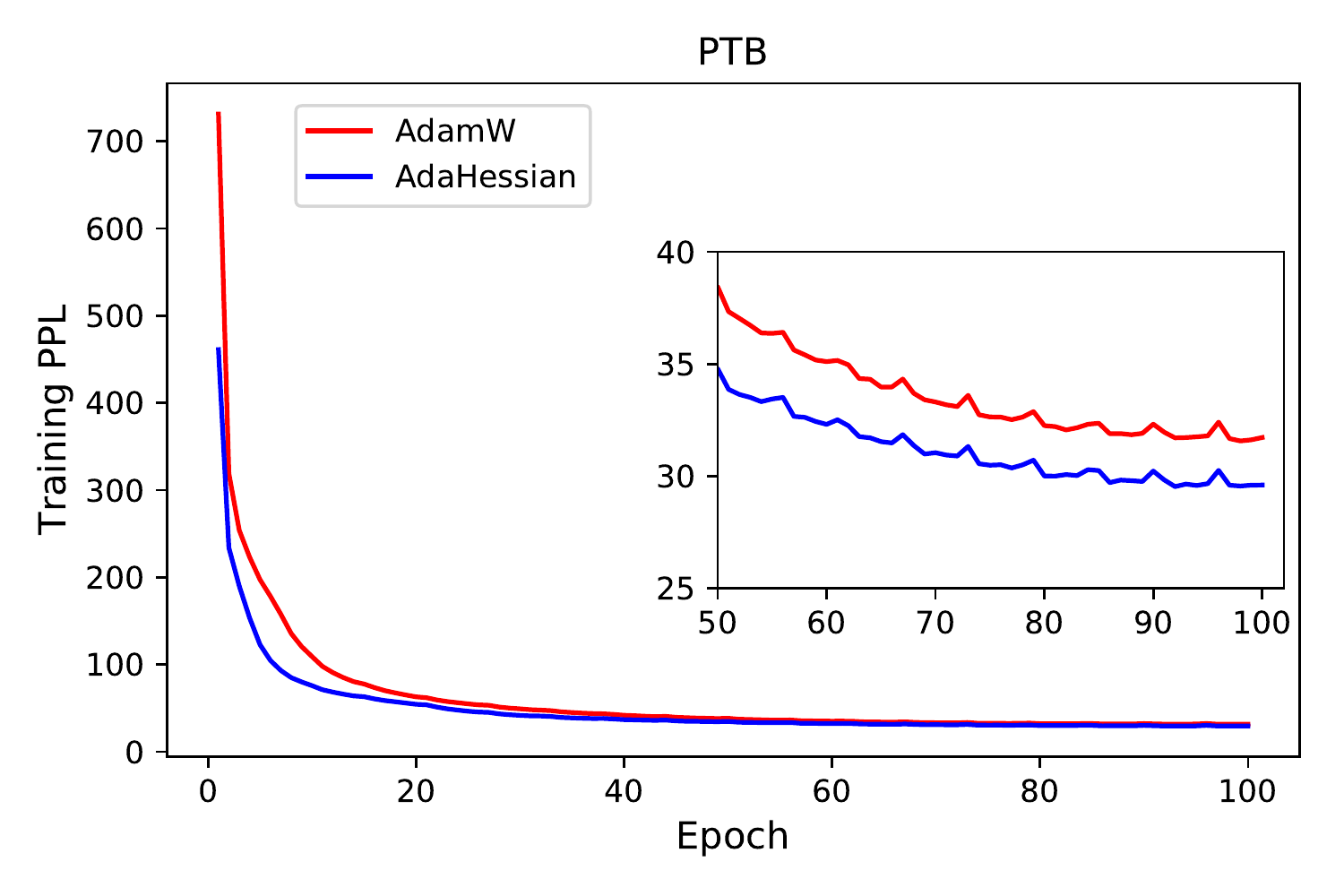}
  \includegraphics[width=.45\linewidth]{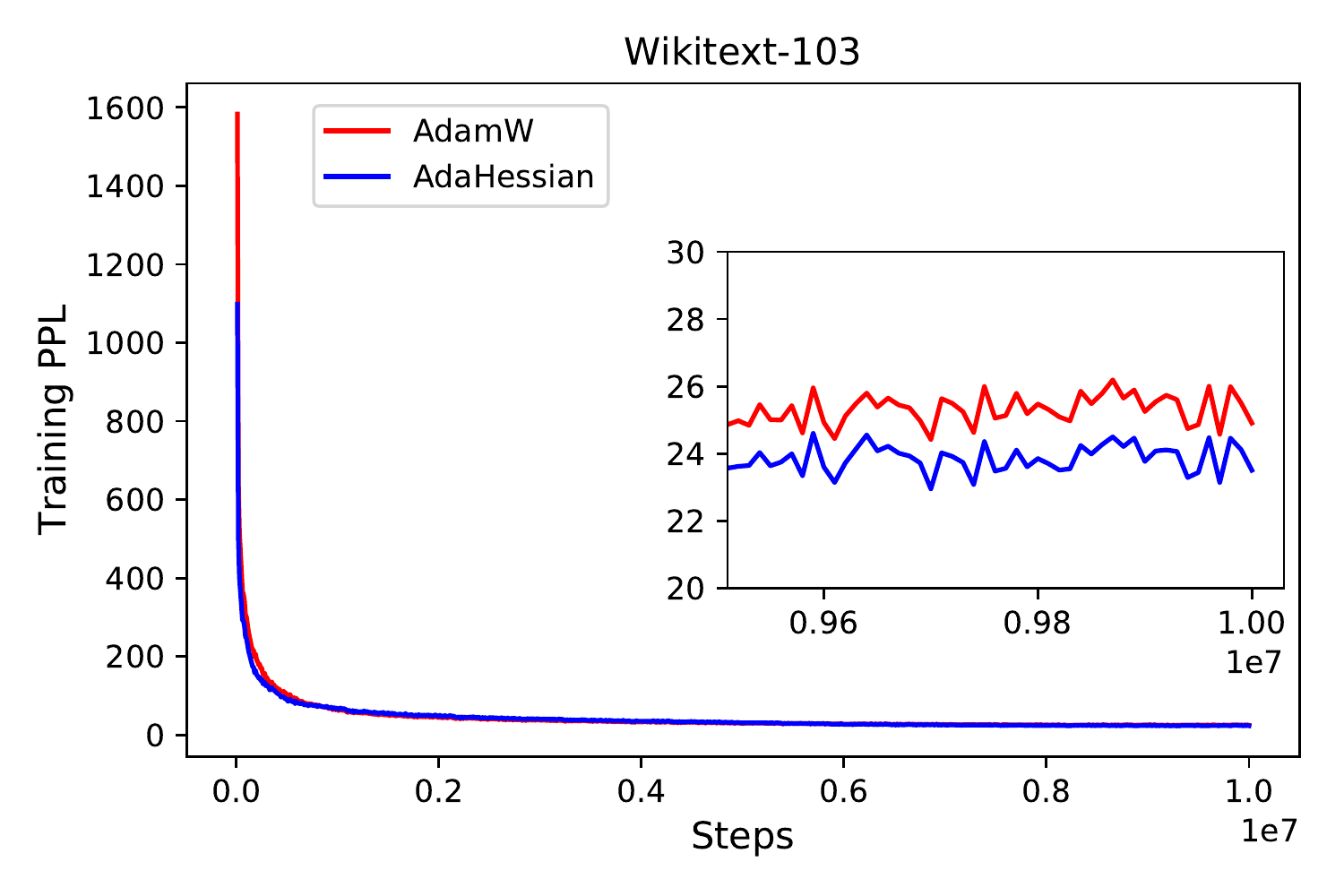}
\end{center}
 \caption{\footnotesize 
 Training PPL curves of \adamw and \OURS for Transformer on PTB and Wikitest-103. 
 The losses of \OURS are consistently lower than \adamw from the beginning of the training.
 \OURS achieves 29.56/23.51 final training perplexity (PPL) on PTB/Wikitext-103 as compared to \adamw (31.72/24.01). Testing results are reported in~\tref{tab:language_modeling}.
 }
\label{fig:ptb_training_curve}
\end{figure*}

\end{document}